%% file: main.tex
\documentclass{article}

\usepackage[preprint]{neurips_2025}

\usepackage{booktabs}
\usepackage{multirow}
\usepackage{multicol}
\usepackage{color, colortbl}
\usepackage{pifont}%
\usepackage{array}
\usepackage{amsmath}
\usepackage{amssymb}
\usepackage{savesym}
\usepackage{paralist}

\savesymbol{todo}
\usepackage{todonotes}
\restoresymbol{t}{todo}

\usepackage[utf8]{inputenc} % allow utf-8 input
\usepackage[T1]{fontenc}    % use 8-bit T1 fonts
\usepackage{hyperref}       % hyperlinks
\usepackage{url}            % simple URL typesetting
\usepackage{booktabs}       % professional-quality tables
\usepackage{amsfonts}       % blackboard math symbols
\usepackage{nicefrac}       % compact symbols for 1/2, etc.
\usepackage{microtype}      % microtypography
\usepackage{xcolor}         % colors
\usepackage{subcaption} % 
\usepackage{graphicx}

\input{preamble.tex}

\title{\mtd{}: Adding Scene-Centric Forecasting \\ Control to Occupancy World Model}
\author{\textbf{Yining Shi$^{1,2}$, Kun Jiang$^{1,2\dagger}$, Qiang Meng$^{3}$, Ke Wang$^{3}$, Jiabao Wang$^{4}$, }\\ \textbf{Wenchao Sun$^{1,2}$,Tuopu Wen$^{1,2}$, Mengmeng Yang$^{1,2}$, Diange Yang$^{1,2\dagger}$ }
\thanks{$^{1}$ School of Vehicle and Mobility, Tsinghua University, Beijing, China. $^{2}$ State Key Laboratory of Intelligent Green Vehicle and Mobility, Beijing, China. $^{3}$ Kargobot Inc. $^{4}$ Nankai University. $\dagger$: Corresponding authors: Diange Yang, Kun Jiang (ydg@mail.tsinghua.edu.cn, jiangkun@mail.tsinghua.edu.cn.)}
}

\begin{document}

\maketitle

% \begin{abstract}
%   The abstract paragraph should be indented \nicefrac{1}{2}~inch (3~picas) on
%   both the left- and right-hand margins. Use 10~point type, with a vertical
%   spacing (leading) of 11~points.  The word \textbf{Abstract} must be centered,
%   bold, and in point size 12. Two line spaces precede the abstract. The abstract
%   must be limited to one paragraph.
% \end{abstract}

\input{sec/0_abstract_b}
\input{sec/1_intro_b}
\input{sec/2_related}

\input{sec/3_method}

\input{sec/4_exp}
\input{sec/5_conclu}

% \begin{ack}

% \end{ack}

\newpage
\small{
\bibliographystyle{plain}
\bibliography{ref}
}
\newpage

%%%%%%%%%%%%%%%%%%%%%%%%%%%%%%%%%%%%%%%%%%%%%%%%%%%%%%%%%%%%

\input{sec/6_appendix}

\end{document}

%% file: preamble.tex
\def\mtd{COME}
\def\fvm{scene-centric prediction branch} 
\def\ctln{ControlNet}
\def\occnus{Occ3D-nuScenes}

\usepackage[capitalize]{cleveref}
\crefname{section}{Sec.}{Secs.}
\Crefname{section}{Section}{Sections}
\crefname{table}{Tab.}{Tabs.}
\Crefname{table}{Table}{Tables}

%% file: sec/0_abstract_b.tex
\begin{abstract}
World models are critical for autonomous driving to simulate environmental dynamics and generate synthetic data.
Existing methods struggle to disentangle ego-vehicle motion (perspective shifts) from scene evolvement (agent interactions), leading to suboptimal predictions.
Instead, we propose to separate environmental changes from ego-motion by leveraging the scene-centric coordinate systems.
In this paper, we introduce \mtd{}: a framework that integrates scene-centric forecasting \underline{C}ontrol into the \underline{O}ccupancy world \underline{M}od\underline{E}l.
Specifically, \mtd{} first generates ego-irrelevant, spatially consistent future features through a \fvm{}, which are then converted into scene condition using a tailored \ctln{}.
These condition features are subsequently injected into the occupancy world model, enabling more accurate and controllable future occupancy predictions.
Experimental results on the nuScenes-Occ3D dataset show that \mtd{} achieves consistent and significant improvements over state-of-the-art (SOTA) methods across diverse configurations, including different input sources (ground-truth, camera-based, fusion-based occupancy) and prediction horizons (3s and 8s).
For example, under the same settings, \mtd{} achieves 26.3\% better mIoU metric than DOME~\cite{dome} and 23.7\% better mIoU metric than UniScene~\cite{uniscene}.
These results highlight the efficacy of disentangled representation learning in enhancing spatio-temporal prediction fidelity for world models.
Code and videos will be available at \url{https://github.com/synsin0/COME}.
\end{abstract}

%% file: sec/1_intro_b.tex
\section{Introduction}\label{sec:intro}
World models are designed to discern the current state of the environment and predict subsequent states based on executed actions. 
%Through simulation of prospective scenarios, world models facilitate a comprehensive understanding of environmental evolution under varying conditions.
This predictive ability of world models not only enables the assessment of decision-making consequences but also facilitates the generation of synthetic data, which serves as a crucial resource for training, testing, and simulation in autonomous systems. 
Consequently, world models have emerged as a focal point of research in the field of autonomous driving, garnering substantial attention from both academia and industry.

\begin{figure}
    \centering
    \includegraphics[width=0.9\textwidth]{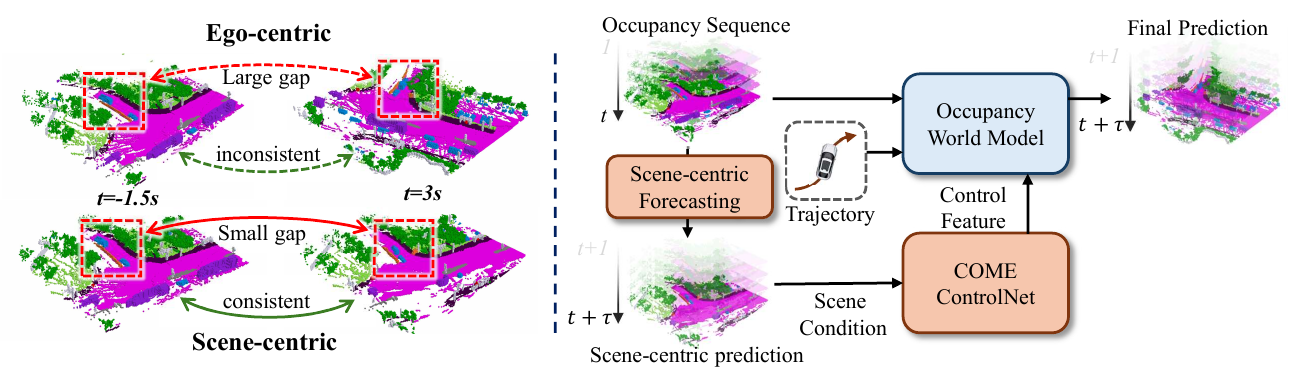}
    \caption{\mtd{} with both scene-centric and ego-centric representation. Compared to ego-centric evolution, scene-centric prediction shows smaller gap in the context of temporal evolution. \mtd{} uses scene-centric prediction results as an important guidance to enhance occupancy world model.
    }
    \label{fig:compare_video_occ}
    % \vspace{-1em}
\end{figure}

The synthetic data generated by world models can take various forms to represent future scenes.
%These include 2D videos that visually depict future environments~\cite{hu2023gaia, uniscene, wang2024drivedreamer, wang2024driving},
These include 2D videos~\cite{hu2023gaia, uniscene, wang2024drivedreamer, wang2024driving},
%3D lidar point clouds that offer a detailed spatial representation of the surroundings~\cite{agro2024uno, khurana2023point, uniscene, yang2024visual, zyrianov2024lidardm},
3D lidar point clouds~\cite{agro2024uno, khurana2023point, uniscene, yang2024visual, zyrianov2024lidardm},
%and 3D occupancy grids that provide a discretized representation of the space occupied by objects~\cite{occworld, occvar, dome,dfit-occworld,occ-llm, uniscene, renderworld, occllama}.
and 3D occupancy grids~\cite{occworld, occvar, dome,dfit-occworld,occ-llm, uniscene, renderworld, occllama}.
Irrespective of the representation format, the changes in the appearance of future scenes predicted by world models are predominantly governed by two key factors:
1) Ego-vehicle motion: The movement of the autonomous vehicle alters its viewing perspective, leading to dynamic changes in perceived spatial features (e.g., perspective shifts, occlusions).
2) Scene evolvement: Natural changes in the environment, such as agent interactions (e.g., pedestrian movements, vehicle collisions) and background updates (e.g., traffic light changes, weather variations).
%The first factor, ego-vehicle motion, pertains to the alterations in the visual and spatial information perceived by an autonomous vehicle as it traverses the environment.
%As the vehicle moves, the perspective from which it observes the world changes, leading to alternations in the model's representation.
%The second factor, scene evolvement, encompasses changes arising from modifications in the background environment and interactions among various agents, such as vehicles and pedestrians.
Notably, these two sources of change exhibit distinct characteristics. In a wide range of real-world scenarios, the background environment may remain relatively stable, with minimal changes over time (see \cref{fig:compare_video_occ}).
Instead, it is the motion of the ego-vehicle and the resulting shifts in the viewing perspective that contribute significantly to the dynamic changes in the world model's representation.

State-of-the-art (SOTA) world models, however, rely on neural networks to implicitly learn the intertwined effects of these factors, often resulting in suboptimal spatial consistency. As a case study, the SOTA model DOME~\cite{dome} achieves an average 27.10 mIoU for 3-second occupancy prediction using ground-truth trajectories. In contrast, when predictions are formulated in a scene-centric coordinate system (where static backgrounds are decoupled from ego motion), a vanilla U-Net achieves 39.12 mIoU—a 44\% improvement. This disparity underscores the critical lack of explicit control over spatial consistency in existing methods, motivating the need for disentangled representation learning.

%However, a critical analysis of state-of-the-art world model methodologies reveals that they predominantly rely on neural networks to implicitly learn the intertwined representation changes associated with both factors simultaneously.
%In this study, we hypothesize that separating these two fundamental components can significantly mitigate the challenges inherent in world model tasks.
%For instance, the state-of-the-art DOME~\cite{dome} can only achieve an average 27.10 mIoU for the 3-seconds prediction task, with ground truth future trajectories.
%In constrast, if we move the prediction into a scene-centric coordinate system, the inherent world model difficulties can be decoupled as the static background environments no longer changes with ego-motion. Under such scene-centric settings, a vanilla U-Net can achieve a 39.12 mIoU, an astonishing 44\% improvement.
%This significant gap highlights the lack of explicit spatial consistency control in existing world models, which motivates us to introduce such control by decoupling environmental dynamics from ego-motion into scene-centric representations.

%%%%%%%%%%%%%%%%%%%
To address this, we propose \mtd{}, a three-stage framework that leverages scene-centric coordinates to separate ego motion from scene dynamics (see \cref{fig:method}):
1) Pose-conditioned generative stage: A diffusion-based model generates future occupancy maps by iteratively denoising latent vectors, using historical occupancy encodings and pose/BEV layouts as conditional inputs.
2) Fixed-view forecasting branch: By transforming past and future frames into a common coordinate system, this stage mitigates ego-motion effects, enabling non-interactive occupancy prediction without explicit scene flow estimation.
3) Global denoising diffusion stage: Cascaded Diffusion Transformer~\cite{DiT} blocks extract spatio-temporal features, with skip connections ensuring cross-layer consistency.
The Model Transfer Design (MTD) acts as the core mechanism, transferring knowledge from fixed-view forecasts to variable-view generation.
Structured as a trainable copy of the generative model's first half, it injects scene-centric condition features into the latter half via skip connections, enhancing generative realism and temporal consistency.

Extensive experiments on the nuScenes-Occ3D dataset validate \mtd{}'s efficacy across diverse settings.
\begin{inparaenum}[a)]
\item Input sources: ground-truth occupancy, camera-based predictions, and fusion-based occupancy inputs;
\item Prediction horizons: short-term (3s) and long-term (8s) forecasts;
\item Trajectory types: ground-truth poses and end-to-end planned trajectories.
\end{inparaenum}

In all configurations, \mtd{} outperforms SOTA baselines by significant margins, with scene-centric settings demonstrating the largest gains. Key contributions include:
\begin{inparaenum}[1)]
\item A divide-and-conquer strategy for occupancy world models, decomposing complex spatio-temporal prediction into ego-motion-agnostic and scene-dynamic components;
\item \mtd{} ControlNet, which leverages scene-centric forecasts as guidance to improve generative fidelity and spatial consistency;
\item Empirical validation of disentangled representation learning, establishing new state-of-the-art performance on a challenging autonomous driving benchmark.
%\item We propose a divide-and-conquer strategy for the occupancy world model, which makes it easier to understand the scene evolution with ego actions.
%\item We propose COME ControlNet, which first uses scene-centric forecasting to understand the evolution of the environment, and then injects it into the generative world model as a guidance to improve the realism and consistency of the generation.
%\item Extensive experiments on nuScenes-Occ3D benchmark under multiple settings show that the proposed \mtd{} significantly enhances the occupancy generation performance to the state-of-the-art.
\end{inparaenum}

%% file: sec/2_related.tex
\section{Related Works}\label{sec:relatedworks}

\subsection{World Model in Autonomous Driving}
%World models, which generate future scenarios based on historical observations from the ego viewpoint or other conditions, are gaining increasing interest in the field of autonomous driving.
Visual world models~\cite{hu2023gaia, wang2024drivedreamer, wang2024driving} leverage 2D video representations, offering great scalability due to the easy accessibility of camera data. 
However, the lack of 3D geometry understanding limits their fidelity in autonomous driving applications.
In contrast, LiDAR-based representation~\cite{agro2024uno, khurana2023point, uniscene, yang2024visual, zyrianov2024lidardm} provides rich geometric comprehension but falls short in semantic-aware generation.
Recently, occupancy representation~\cite{occworld, occvar, dome, dfit-occworld, occ-llm, uniscene, renderworld, occllama} have emerged as a compelling alternative for world modeling, due to their ability to encode both geometric and semantic information simultaneously. Occupancy as intermediate representation also serves as strong geometric prior condition for downstream generation of driving videos\cite{drivingsphere} and LiDAR point clouds\cite{uniscene}.  

Some occupancy-based approaches~\cite{drive-occworld, dfit-occworld} leverage occupancy flow to forecast future scenarios.
While achieving promising results, they often depend on additional annotations and struggle to produce imaginative predictions.
Recent works have shifted toward generative frameworks~\cite{occworld,dome,occ-llm,occllama}: for example, OccWorld~\cite{occworld} employs a two-stage pipeline, first tokenizing occupancy with a VQ-VAE~\cite{vqvae} and then predicting ego motion and scene evolution autoregressively.
Inspired by the success of large language models, some works~\cite{occ-llm,occllama} further enhance the model interpretability during generation. 
Recent methods~\cite{dome,uniscene} employ diffusion transformers (DiTs), demonstrating strong generative capabilities for world modeling.
In this work, we similarly adopt the DiT paradigm while introducing control features inspired by \ctln{}~\cite{controlnet} — a pioneering framework for conditional generation in 2D images using multi-modal inputs (e.g., depth maps, edges, or sketches). 
By incorporating explicit control signals, our method achieves superior spatial consistency and prediction accuracy, establishing new state-of-the-art performance for occupancy world models.

\subsection{4D Occupancy Forecasting}
Occupancy world models and 4D occupancy forecasting methods share a common paradigm of predicting future occupancy states. 
The key distinction is that occupancy forecasting typically targets short-term scene evolution and does not involve predicting future ego trajectories. 
Occ4Cast~\cite{occ4cast} and Cam4DOcc~\cite{cam4docc} establish LiDAR-based and camera-based benchmarks, respectively, and propose baseline models using temporal recurrent networks such as ConvLSTM~\cite{convlstm}. 
To address the high cost of 4D occupancy annotation, 4D-Occ-Forecasting~\cite{4doccforecasting} leverages future point clouds as proxies for future occupancy and employs differentiable depth rendering for self-supervised learning. 
Building upon this, Vidar~\cite{vidar} and UnO~\cite{uno} introduce latent rendering and continuous 4D fields, respectively, to further enhance self-supervised learning for 4D occupancy prediction and related downstream tasks.
In our work, the scene-centric prediction module adopts a similar paradigm to 4D occupancy forecasting - predicting future occupancy while explicitly factoring out the influence of ego trajectory. 
This design establishes a spatially consistent control prior that effectively guides the learning process.% of our generative model.

%% file: sec/3_method.tex
\section{Methodology}\label{sec:method}

\begin{figure}[t]
    \centering
    \includegraphics[width=1.0\textwidth]{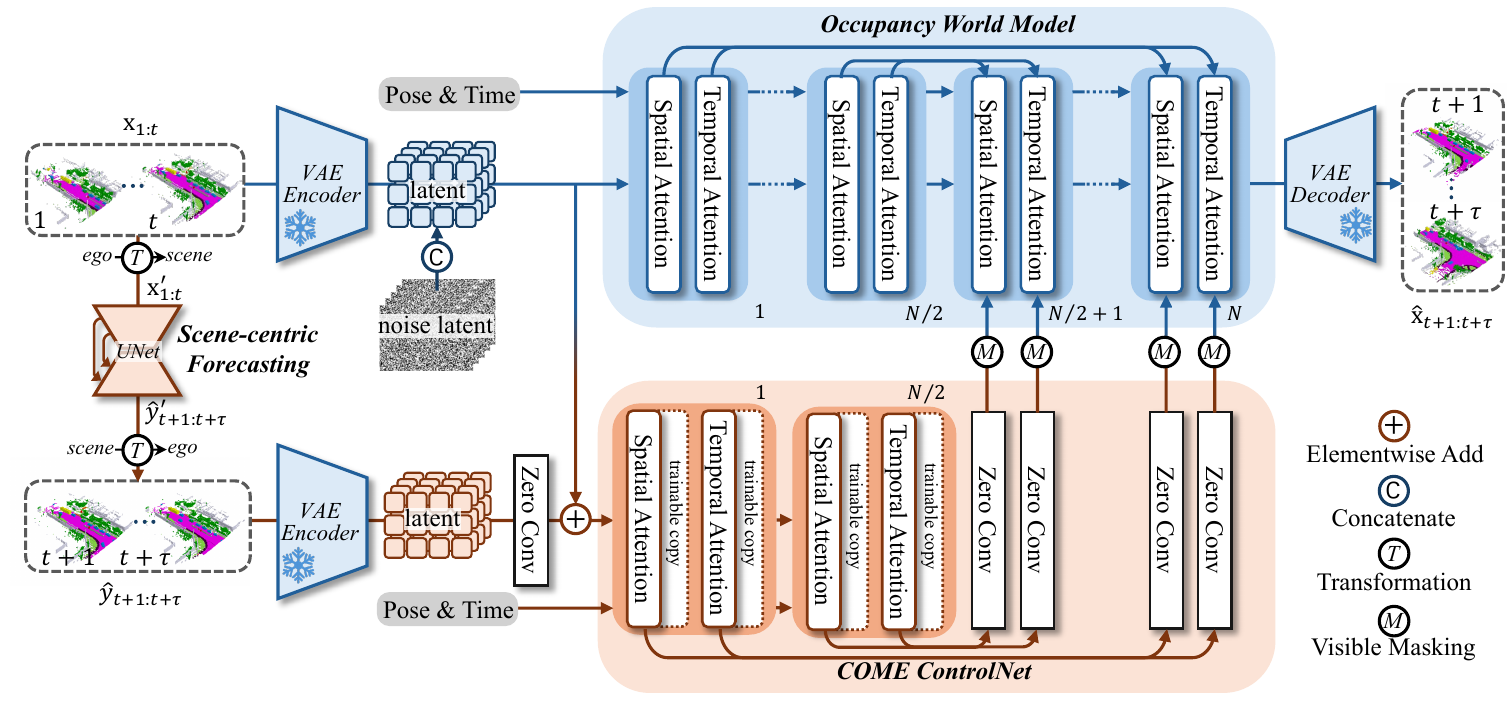}
    \caption{
        The proposed \mtd{} framework comprises three main modules:
        (1) an \textbf{Occupancy World Model} that predicts future occupancy using historical observations and other inputs (e.g., poses, time steps, BEV layouts);
        (2) a \textbf{Scene-centric Forecasting Module} that produces spatially consistent scene predictions by eliminating the influence of ego motion; and
        (3) the \textbf{\mtd{} \ctln{}} which converts the scene conditions from the forecasting module into control features that are subsequently injected into the world model for controllable and geometrically coherent occupancy generation.
        }        
    \label{fig:method}
    % \vspace{-1em}
\end{figure}

This section presents our \mtd{}, a framework that integrates scene-centric forecasting \underline{C}ontrol into the \underline{O}ccupancy world \underline{M}od\underline{E}l.
We detail the three main components of \mtd{} (illustrated in \cref{fig:method}) in \cref{sec:method_wm,sec:method_sfm,sec:method_ctrl}.
Finally, \cref{sec:method_loss} describes the training objectives and the overall training pipeline.
% Our approach consists of three core components:
% (1) a \textbf{Occupancy World Model} (\cref{sec:method_wm}) that serves as the foundation for the task,
% (2) a \textbf{Scene-centric Forecasting Module} (\cref{sec:method_sfm}) that generates ego-irrelevant future control priors,
% and (3) the \textbf{\mtd{} \ctln{}} (\cref{sec:method_ctrl}) that injects scene-centric control features to improve spatial consistency of the world model.
% Finally, \cref{sec:method_loss} describes training objectives and pipeline in detail.

\subsection{Occupancy World Model}\label{sec:method_wm}
%Before introducing our improvements, 
We first describe our modified baseline world model, which is capable of performing the generation task independently. 
Following DOME~\cite{dome}, our model leverages diffusion Transformers (DiTs) for the superior fine-grained and imaginative generation compared to auto-regressive counterparts~\cite{occworld,dome,occ-llm,occllama}. 
Given historical observations $\mathbf{x}_{1:t}$, ego-vehicle states $p_{1:t}$ and other possible inputs (e.g., BEV layouts), the occupancy world model predicts future occupancy $\mathbf{\hat{x}}_{t+1:t+\tau}$  while optionally forecasting future trajectories $\hat{p}_{t+1:t+\tau}$.
We treat future trajectories — whether ground-truth or predicted by a planning module — as available inputs for the generative model for two key reasons:
(1) It ensures comparability with existing methods, since some directly generate with ground-truth trajectories while others predict by themselves.
(2) It enables trajectory-conditioned generation, which is essential for world models to produce diverse and controllable outputs.

Our model architecture comprises two principal components.
The first is a collection of input encoders that transform historical observations into compact representations for efficient processing during the diffusion stage.
We adopt the trajectory encoder from DOME~\cite{dome} to encode trajectories, and apply a max-pool layer for BEV layouts when available.
For the occupancy data, we utilize a Occupancy Variational Auto-Encoder (Occ-VAE) consisting of an encoder network $q_{\phi}(\mathbf{z}|\mathbf{x})$ and a decoder network $p_{\theta}(\mathbf{x}|\mathbf{z})$.
Given input occupancy $\mathbf{x}\in \mathcal{R}^{H\times W \times D}$, $q_{\phi}(\mathbf{z}|\mathbf{x})$ encodes it into a continuous latent variable $\mathbf{z}\sim q_{\phi}(\mathbf{z}|\mathbf{x})$.
Conversely, $p_{\theta}(\mathbf{x}|\mathbf{z})$ is able to reconstruct the occupancy from the latent variable.
This design enables the Occ-VAE to compress high-dimensional occupancy data into a compact latent space, facilitating more efficient processing in the generative model.

The second component is the spatial-temporal Diffusion Transformer modified from the DOME~\cite{dome}, which consists of alternating stacked spatial and temporal layers. 
To reserve interface for later feature conditioning, we introduce skip connections between every early and late layer pair, similar to the architectural designs in UNet~\cite{unet} and HunyuanDiT~\cite{hunyuandit}. For example, the last spatial block takes the first spatial block as a skip input. For each block in the second half, the input and the skip are concatenated in channel dimension and undergo a simple MLP for feature fusion.

The model processes latent features encoded from occupancy data, along with optional inputs, eg BEV layouts if available.
Trajectories are converted into trajectory condition and injected into each spatial-temporal block, enabling trajectory-aware generation.
In the end, the outputs are decoded through $q_{\phi}(\mathbf{z}|\mathbf{x})$ into occupancy predictions conditional to various future time steps and ego poses. 

\subsection{Scene-centric Forecasting Module}\label{sec:method_sfm}
Future occupancy prediction represents a complex spatial-temporal modeling challenge that requires reasoning about two key factors: (1) the natural evolution of the scene (e.g., moving objects, infrastructure changes), and (2) the planned trajectory of the ego-vehicle. 
Our foundation world model described in \cref{sec:method_wm} treats the future trajectory as a condition but has limited understanding of the inherent nature of scene evolution, leading to spatially inconsistent scene evolution in the generated occupancy — a limitation also confirmed in later experiments. 
To address this, we propose a scene-centric forecasting module that produces more coherent scene evolution and generates scene-conditioned inputs for the world model.

In practice, the scene exhibits relatively smaller changes when represented in a unified coordinate frame compared to an ego-centric occupancy representation. 
In \occnus{}, static voxels account for over 92.7\% of all occupied voxels, suggesting that forecasting the scene in a shared coordinate system simplifies the task without requiring complex designs. 
In our implementation, we find that a simple UNet~\cite{unet} suffices for high-quality future predictions, where the skip connections effectively preserve static voxels, while the multi-scale structure captures dynamic elements.

Given historical occupancy $\mathbf{x}_{1:t}$ and ego-vehicle states $p_{1:t}$,  we eliminate the influence of ego motion by transforming each occupancy frame into the current frame  through rigid body transformations. 
Specifically, we initialize an empty occupancy grid aligned with the ego pose $p_t$, which is then populated by transforming the voxels in each $\mathbf{x}_i$ into the current frame using nearest-neighbor search and grid sampling, resulting in $\mathbf{x}'_i = T(\mathbf{x}_i, p_i, p_t)$.
The transformed sequence $\mathbf{x}'_{1:t}$ is reshaped into a 3D tensor of dimensions $Dt \times H \times W $, allowing efficient processing with 2D convolutions across time steps, rather than using computationally expensive 3D convolutions.
This tensor is processed by the network to produce an output of size $D\tau \times H \times W$, which is then decoded into occupancy predictions $\mathbf{\hat{y}'}_{t+1:t+\tau}$.
Finally, we transform these predictions into their respective future time steps and encode them via the Occ-VAE encoder, producing the scene condition $\mathbf{z}_i = q_{\phi}(\mathbf{z}|\mathbf{\hat{y}}_i)$ with $\mathbf{\hat{y}}_{i} = T(\mathbf{\hat y'}_{i}, p_t, p_{i}), i \in\{ t+1, t+2, \cdots, t+\tau$\}.

% Given historical occupancy $\mathbf{x}_{1:t}$ and ego-vehicle states $p_{1:t}$, we transform each past occupancy $\mathbf{x}_i$ into current frame's voxels, yielding $\mathbf{x}'_i = T(\mathbf{x}_i, p_i, p_t)$.
% \qtext{to be polished later}
% The $\mathbf{x}'_{1:t}$  is reshaped into a tensor of dimensions $B \times Dt_{src} \times L \times W$, enabling efficient processing with 2D convolutions rather than computational-expensive 3D ones. $B$ is the batch size, $D$ is the class channel multiplying with height channel, $t_{src}$ is the source sequence length. The output of the UNet module is a tensor of dimensions $B \times Ct_{tgt} \times L \times W$, where $t_{tgt}$ is the target sequence length and $C$ is the base channel. After UNet, the feature is reshaped to $Bt_{tgt} \times C \times L \times W$. Several Conv2D blocks further decode the spatial features of each frame. The base channel $C$ is transformed into $H \times N_{c}$ with the last Conv2D blcok, where $H$ is the height of the grid size and $N_c$ is the number of semantic categories. Finally, the network outputs a $B \times t_{tgt} \times H \times N_{c} \times L \times W$, which is reshaped into predictions $\mathbf{\hat x}'_{t+1:t+\tau}$.
% Finally, we transform these predictions into their respective future time steps and encode them via the Occ-VAE encoder, producing the scene condition $\mathbf{z}_{t+1:t+\tau} = q_{\phi}(\mathbf{z}|T(\mathbf{\hat x}'_{t+1:t+\tau}, p_t, p_{t+1:t+\tau}))$.

\subsection{\mtd{} \ctln{}}\label{sec:method_ctrl}

The \mtd{} \ctln{}, which encodes scene conditions $\{\mathbf{z}_i\}_{i=1}^{\tau} \in \mathbb{R}^{C \times H_1\times W_1}$ into implicit control features, consists of $N$ blocks inspired by HunyuanDiT\cite{hunyuandit}.
As shown in \cref{fig:method}, the first $N/2$ blocks are trainable copies of the corresponding spatial-temporal blocks from the occupancy world model, while the remaining blocks are the zero convolution layers.
Each of these layers outputs control feature $\{\mathbf{c}^n_i\}_{i=1}^{\tau} \in \mathbb{R}^{C \times H_1 \times W_1}$ for $n=N/2+1, \cdots, N$, where $\tau$ denotes the number of future frames, and $C, H_1, W_1$ represent the channel depth, height, and width dimensions, respectively.

However, we observe that $\{\mathbf{c}^n_i\}_{i=1}^{\tau}$ is not directly suitable as a control prior for world model. 
This limitation arises because the forecasting module, due to its simple structure, lacks sufficient capacity for imagining future states. 
Consequently, its predictions in historically unobserved regions may bring noise, degrading the world model's generative performance — a finding corroborated by late experiments. 
Thus, filtering unreliable features from control features is crucial for robust generation.

To address this, we propose a visibility-aware masking strategy based on 3D spatial relationships. 
For each future timestamp $i\in \{t+1, t+2, \cdots, t+\tau\}$, we first trace the root source of the control feature $\mathbf{c}^n_i$, namely the $\mathbf{\hat y}_i$ predicted by the scene-centric forecasting module.
We then construct a binary voxel mask $\mathbf{m}_{i} \in \mathbb{R}^{D \times H\times W}$, where a value of 1 indicates that the corresponding voxel center in $\mathbf{\hat y}_i$ is observable in historical occupancy data, and 0 otherwise. 
This allows us to derive an invisibility mask $\mathbf{M}_{i} \in \mathbb{R}^{H_1\times W_1}$ that  quantifies the reliability of each spatial feature in $\mathbf{c}_i^n$:
\begin{equation}
    \small
    \mathbf{M}_i(h, w) = \mathbb{I}
\left(\frac{1}{D\cdot \delta_h \cdot \delta_w}\sum_{d=1}^{D}\sum_{u=h\cdot \delta_h}^{(h+1)\cdot \delta_h} \sum_{v=w\cdot \delta_w}^{(w+1)\cdot \delta_w} \mathbf{m_i}(d, u, v)  < \varepsilon \right),
\end{equation}
where $\delta_h = H/H_1, \delta_w = W/W_1$ and $\varepsilon$ is a pre-defined threshold. In practice we set $\varepsilon=0.5$.
The $\mathbb{I}$ is an indicator function to check whether proportion of historically observed voxels within each pillar (corresponding to feature location (h, w) in $\mathbf{c}_i^n$) is large enough.
Then, we suppress unreliable features via element-wise multiplication: $\mathbf{c}^n_i \leftarrow \mathbf{c}^n_i \odot \mathbf{M}_i$.
Finally, these refined control features are injected into the world model through residual addition at corresponding layers, ensuring robust and controllable generation. The control features are added to skip features and concatenated together to the input features for each blocks in the second half of the world model.

\subsection{Training Pipelines}\label{sec:method_loss}

Inspired by \ctln{}~\cite{controlnet},  we employ a multi-stage training pipeline as 
(1) In \textbf{stage 1}, the occupancy world model is trained with configurations introduced in DOME~\cite{dome}. 
This stage establishes a strong foundational occupancy generation ability.
(2) In \textbf{stage 2}, we train the UNet-based forecasting module using a simple cross-entropy loss.
(3) In \textbf{stage 3}, we freeze all other modules and exclusively train the parameters of \ctln{}. 
Our multi-stage training strategy not only optimizes training efficiency but also ensures controllable generation, as demonstrated in our ablation studies.

%% file: sec/4_exp.tex
\section{Experiments}\label{sec:exp}

%\mtd{} consistently surpasses baseline approaches across a variety of settings.%, demonstrating the versatility and robustness of our method in different conditions.
We elaborate our experimental settings in \cref{sec:exp_setup}, the quantitative and qualitative results of the proposed \mtd{} framework in \cref{sec:exp_quant,sec:exp_quali}, 
and extensive ablation studies in \cref{sec:exp_ab}.

\subsection{Experimental Setup}\label{sec:exp_setup}

\textbf{Dataset and metrics}
All experiments are conducted on the widely used \occnus{}\cite{occ3d} benchmark, which offers 3D occupancy labels for 18 categories 
% (1 free space class, 16 semantic classes, and a generic obstacle class)
 based on the large-scale nuScenes\cite{nuscenes} dataset.
The annotations cover a spatial range of [-40m, -40m, -3.2m, 40m, 40m, 3.2m] with a voxel resolution of [0.4m, 0.4m, 0.4m], resulting in a $200\times 200 \times 16$ voxel grid per frame.
The dataset is split into 700 training, 150 validation, and 150 test driving sequences, each lasting 20 seconds.
We adopt geometric Intersection over Union (IoU) and semantic mean Intersection over Union (mIoU) as evaluation metrics. Results are reported at each future timestamp, along with the average performance over all timestamps on the validation set, in line with previous works.

\textbf{Implementation details.}\label{sec:imp_detail}
Unless otherwise specified, our world model predicts future occupancy over a 3-second horizon at 2 frames per second, conditioned on 4 frames of historical occupancy data, following the protocol established in OccWorld~\cite{occworld}.
As outlined earlier, the \mtd{} framework is trained in multiple stages:
(1) \textit{Diffusion-based World Model.} 
We adopt the pre-trained Occ-VAE from DOME~\cite{dome} and train the diffusion-based world model for 2000 epochs with a total batch size of 128 and a learning rate of 2e-4.
(2) \textit{Scene-centric Forecasting Module.} 
This module is trained for 12 epochs using a total batch size of 32 and the CBGS resampling strategy~\cite{cbgs}.
(3) \textit{\mtd{} \ctln{}.} 
This component is trained for 1000 epochs with a total batch size of 64. 
All models are trained on 4 H20 GPUs and use a learning rate of 1e-4 is not stated specifically.
Please refer to the supplementary materials (\cref{app:imp_detail}) for additional implementation details, including network architectures and statistics, training hyperparameters, and planning trajectory configurations.

\subsection{Quantitative Evaluation}\label{sec:exp_quant}

\begin{table}[t]
    \newcommand{\graycell}{\cellcolor{gray!30}}
    \setlength{\tabcolsep}{0.006\linewidth}
    \caption{
        \textbf{4D occupancy generation performance under various settings.}
        Each setting varies in terms of input modality and whether the ego trajectory (ego traj.) is predicted (Pred.) by a planning module or provided as ground truth (GT). 
        % Most settings use 4-frame historical input except that for comparison with UniScene-Fore, both methods use 2-frame historical input.
        ``Avg." indicates the average performance across 1s, 2s, and 3s horizons. The best performance in each setting is highlighted in bold.
    }\label{tab:compare_sota}
    \centering
    \small
    \begin{tabular}{l|cc|cccc|cccc}
        \toprule            
        \multirow{2}{*}{Method} & \multirow{2}{*}{Input} & \multirow{2}{*}{Ego traj.} &
        \multicolumn{4}{c|}{mIoU (\%) $\uparrow$} & 
        \multicolumn{4}{c}{IoU (\%) $\uparrow$} \\
         & & & 1s & 2s & 3s & \graycell  Avg. & 1s & 2s & 3s & \graycell Avg. \\
        \midrule
        OccWorld-D~\cite{occworld} &  Camera & Pred. & 11.55 & 8.10 & 6.22 & \graycell 8.62  & 18.90 & 16.26 & 14.43 & \graycell 16.53 \\
        OccWorld-T~\cite{occworld} &  Camera & Pred.  & 4.68 & 3.36 & 2.63 & \graycell 3.56 & 9.32 & 8.23 & 7.47 & \graycell 8.34 \\
        OccWorld-S~\cite{occworld} &  Camera & Pred. & 0.28 & 0.26 & 0.24 & \graycell 0.26  & 5.05 & 5.01 & 4.95 & \graycell 5.00 \\
        OccWorld-F~\cite{occworld} &  Camera & Pred.  &  8.03  & 6.91   &  3.54  &  \graycell 6.16  &     23.62   &  18.13  &  15.22  &  \graycell 18.99    \\
        OccLLaMA~\cite{occllama} &  Camera & Pred. &  10.34  &  8.66  &  6.98  &  \graycell  8.66     &  25.81  &  23.19  &   19.97 &  \graycell  22.99  \\
        OccVAR~\cite{occvar} &  Camera & Pred. &    17.17 &10.38 &7.82 &\graycell 11.79 & 27.60 &25.14 &20.33 & \graycell 24.35  \\
        DFIT-OccWorld~\cite{dfit-occworld} &  Camera & Pred. & 13.38 & 10.16 & 7.96 & \graycell{10.50}  & 19.18 & 16.85 & 15.02 & \graycell{17.02} \\
        Occ-LLM-S~\cite{occ-llm} &  Camera & Pred. & 11.28 & 10.21 & 9.13 & \graycell 10.21  & 27.11  & 24.07 & 20.19 &\graycell 23.79 \\
        RenderWorld-S~\cite{renderworld} &  Camera & Pred. & 2.83  & 2.55 & 2.37 &\graycell 2.58  & 14.61 & 13.61 & 12.98 & \graycell 13.73  \\
        \mtd{ (Ours)} &  Camera & Pred. & \textbf{25.57}  & \textbf{18.35} & \textbf{13.41} & \graycell {\textbf{19.11}}  & \textbf{45.36} & \textbf{37.06} & \textbf{30.46} &\graycell {\textbf{37.63}} \\

% 05/14 14:23:14 - mmengine - INFO - Current val iou is [52.73471474647522, 45.35638391971588, 41.19707643985748, 37.0556116104126, 33.86086821556091, 30.46341836452484] while the best val iou is [52.73471474647522, 45.35638391971588, 41.19707643985748, 37.0556116104126, 33.86086821556091, 30.46341836452484]
% 05/14 14:23:14 - mmengine - INFO - Current val miou is [32.65568136292345, 25.573868194923683, 21.789052819504455, 18.346204312846943, 15.863750940736601, 13.407210249672918] while the best val miou is [32.65568136292345, 25.573868194923683, 21.789052819504455, 18.346204312846943, 15.863750940736601, 13.407210249672918]
% 05/14 14:23:14 - mmengine - INFO - avg val iou is 37.62513796488444
% 05/14 14:23:14 - mmengine - INFO - avg val miou is 19.10909425248118

        \midrule
        DOME~\cite{dome} & Camera & GT &   24.12  &  17.41  &  13.24  &  \graycell 18.25  & 35.18        &  27.90  &  23.44 &  \graycell 28.84  \\
        \mtd{ (Ours)} & Camera & GT & \textbf{26.56}  & \textbf{21.73} & \textbf{18.49} & \graycell {\textbf{22.26}}  & \textbf{48.08} & \textbf{43.84} & \textbf{40.28} &\graycell {\textbf{44.07}} \\

% 05/14 14:23:14 - mmengine - INFO - Current val iou is [51.325541734695435, 48.07914197444916, 45.79783082008362, 43.84497404098511, 42.43859946727753, 40.27961790561676] while the best val iou is [51.325541734695435, 48.07914197444916, 45.79783082008362, 43.84497404098511, 42.43859946727753, 40.27961790561676]
% 05/14 14:23:14 - mmengine - INFO - Current val miou is [30.49533494255122, 26.561552680590573, 23.994348965146962, 21.726973617778107, 20.11303544482764, 18.49478965296465] while the best val miou is [30.49533494255122, 26.561552680590573, 23.994348965146962, 21.726973617778107, 20.11303544482764, 18.49478965296465]
% 05/14 14:23:14 - mmengine - INFO - avg val iou is 44.067911307017006
% 05/14 14:23:14 - mmengine - INFO - avg val miou is 22.261105317111106

        \midrule
        Copy\&Paste~\cite{occworld} & 3D-Occ & Pred. &  14.91 & 10.54 & 8.52 & \graycell 11.33 & 24.47 & 19.77 & 17.31 & \graycell 20.52 \\ 
        OccWorld~\cite{occworld} & 3D-Occ & Pred. & 25.78 & 15.14 & 10.51 & \graycell 17.14  & 34.63 & 25.07 & 20.18 & \graycell 26.63 \\
        OccLLaMA~\cite{occllama}  & 3D-Occ & Pred. &   25.05  &   19.49  &  15.26  &  \graycell 19.93    &  34.56  &  28.53  &  24.41  &   \graycell 29.17 \\
        OccVAR~\cite{occvar} & 3D-Occ & Pred. & 27.96 & \textbf{21.75} & 16.47 & \graycell 22.06 & 38.73 &29.50 & 24.86 & \graycell 31.03 \\
        RenderWorld~\cite{renderworld} & 3D-Occ & Pred. & 28.69 & 18.89 & 14.83  & \graycell 20.80  & 37.74 & 28.41 & 24.08 & \graycell 30.08 \\
        Occ-LLM~\cite{occ-llm} & 3D-Occ & Pred. & 24.02 & 21.65 & \textbf{17.29} & \graycell 20.99   & 36.65 & \textbf{32.14} & \textbf{28.77}  & \graycell \textbf{32.52}  \\
        DFIT-OccWorld~\cite{dfit-occworld} & 3D-Occ & Pred. & \textbf{31.68} & 21.29 & 15.18 & \graycell{\textbf{22.71}}  & \textbf{40.28}
        & 31.24 & 25.29 & \graycell{32.27} \\
        \mtd{ (Ours)} & 3D-Occ & Pred. & 30.57  & 19.91 & 13.38 & \graycell {21.29}  & 36.96 & 28.26 & 21.86 &\graycell {29.03} \\

% 05/14 14:09:05 - mmengine - INFO - Current val iou is [45.3999787569046, 36.96238398551941, 32.511648535728455, 28.256016969680786, 24.85816776752472, 21.864037215709686] while the best val iou is [45.3999787569046, 36.96238398551941, 32.511648535728455, 28.256016969680786, 24.85816776752472, 21.864037215709686]
% 05/14 14:09:05 - mmengine - INFO - Current val miou is [41.0660817342646, 30.569861829280853, 24.729320319259866, 19.90714266019709, 16.29400394637795, 13.382023824926684] while the best val miou is [41.0660817342646, 30.569861829280853, 24.729320319259866, 19.90714266019709, 16.29400394637795, 13.382023824926684]
% 05/14 14:09:05 - mmengine - INFO - avg val iou is 29.027479390303295
% 05/14 14:09:05 - mmengine - INFO - avg val miou is 21.28634277146821

        \midrule
        DOME~\cite{dome} & 3D-Occ & GT & 35.11 & 25.89 & 20.29 & \graycell 27.10    & 43.99 & 35.36 & 29.74 &\graycell 36.36 \\
        % &  &  \mtd{-O-MV (Ours)} &  37.70 & 27.29 & 22.62 & \graycell  29.21  & 44.82 & 36.01 & 30.79 &\graycell 37.20 \\
        % &  &  \mtd{-O  (Ours)}  &  \textbf{38.33} & \textbf{35.05} & \textbf{29.90} & \graycell {\textbf{36.19}}  & \textbf{51.67} & \textbf{45.32} & \textbf{40.90} &\graycell {\textbf{45.97}} \\
        \mtd{ (Ours)} & 3D-Occ & GT & \textbf{42.75} & \textbf{32.97} & \textbf{26.98} & \graycell  \textbf{34.23}  & \textbf{50.57} & \textbf{43.47} & \textbf{38.36} &\graycell \textbf{44.13} \\

        \midrule
        UniScene-Fore~\cite{uniscene} & 3D-Occ(2f),Box,Map   &  GT &   35.37 & 29.59 & 25.08 & \graycell {31.76}    & 38.34 & 32.70 & 29.09 &\graycell 34.84 \\
        % &  & \mtd{-MV  (Ours)}  & 41.91  & 33.71 & 27.43 & \graycell {34.35}  & 46.05 & 39.75 & 34.35 &\graycell {40.05} \\
        % &  & \mtd{(Ours)} & \textbf{45.26}  & \textbf{38.30} & \textbf{33.05} & \graycell {\textbf{38.87}}  & \textbf{51.79} & \textbf{46.20} & \textbf{41.53} &\graycell {\textbf{46.50}} \\
        \mtd{  (Ours)}  & 3D-Occ(2f),Box,Map   &  GT & \textbf{45.98}  & \textbf{38.57} & \textbf{33.28} & \graycell \textbf{39.28}  & \textbf{52.11} & \textbf{46.73} & \textbf{42.65} &\graycell \textbf{47.16} \\

% 05/14 13:57:40 - mmengine - INFO - Current val iou is [56.487977504730225, 52.11188793182373, 48.90182912349701, 46.7320591211319, 44.91143822669983, 42.650291323661804] while the best val iou is [56.487977504730225, 52.11188793182373, 48.90182912349701, 46.7320591211319, 44.91143822669983, 42.650291323661804]
% 05/14 13:57:40 - mmengine - INFO - Current val miou is [52.20244228839874, 45.984514846521265, 41.62844033802257, 38.57071908081279, 36.08140480868956, 33.27648946467568] while the best val miou is [52.20244228839874, 45.984514846521265, 41.62844033802257, 38.57071908081279, 36.08140480868956, 33.27648946467568]
% 05/14 13:57:40 - mmengine - INFO - avg val iou is 47.164746125539146
% 05/14 13:57:40 - mmengine - INFO - avg val miou is 39.277241130669914

        \bottomrule
    \end{tabular}
    % \vspace{-2em}
\end{table}

\textbf{Main results.}
In the occupancy world modeling domain, existing methods differ significantly in their experimental setups. To enable a fair and comprehensive comparison, we adapt our proposed \mtd{} framework to match these various settings and report the main results in \cref{tab:compare_sota}.

% — particularly in the input modality used and whether ego trajectories are predicted by a planning module or provided as ground truth.

The first setting uses camera images as input and predicted ego trajectories from a planning module. 
In this case, we employ a modified BEVDet~\cite{bevdet} to convert camera inputs into occupancy predictions.
Among prior works, OccVAR previously achieved the best performance with an average mIoU of 11.79 and an average IoU of 24.35.
In contrast, our \mtd{} achieve 19.11 mIoU and 37.63 IoU, significantly surpassing the previous best results by 62.1\% and 54.5\%, respectively.

When ground-truth ego trajectories are used with the same camera input, our method further improves performance to 22.26 mIoU and 44.07 IoU, with gains of 3.15 and 6.44, respectively.
In this setting, our method also outperforms the state-of-the-art DOME by a significant margin 22.0\% in mIoU and 34.6\% in IoU, highlighting the robustness of \mtd{}.
% This is expected, as ground-truth trajectories eliminate the planning errors and provide more reliable trajectory conditioning.

In the third setting, we adopt ground-truth occupancy inputs but use predicted ego trajectories.
Here, the best prior mIoU (22.71) is achieved by DFIT-OccWorld, while the highest IoU (32.52) is obtained by Occ-LLM.
It is noteworthy that both methods incorporate additional cues such as occupancy flow or language information into their frameworks.
In comparison, our \mtd{}, despite relying solely on trajectory conditioning without such external information, still achieves competitive results of 21.29 mIoU and 29.03 IoU.
We hypothesize that the slightly lower performance is due to \mtd{}'s strong dependency on trajectory input: since it generates occupancy strictly conditioned on predicted trajectories, it may be more sensitive to planning errors. We further validate the hypothesis in the ablation study. 
Nevertheless, our method remains among the top-performing approaches, demonstrating its effectiveness even in this challenging setup.

Under the configuration with both ground-truth occupancy and ground-truth trajectories, \mtd{} achieves 34.23 mIoU and 44.13 IoU, showing strong performance when free from upstream prediction errors. COME outperforms the state-of-the-art DOME by 26.3\% in mIoU and 21.3\% in IoU.
Furthermore, with BEV layouts, \mtd{}'s performance increases further to 39.28 mIoU and 47.16 IoU. COME outperforms the state-of-the-art UniScene-Fore by 23.7\% in mIoU and 35.4\% in IoU.
% These results confirm that incorporating richer spatial and semantic information can further enhance 4D occupancy generation.

We leave two results in the supplementary material (\cref{sec:more_results}). \cref{tab:compare_long} demonstrates COME achieves better performance than DOME during all timestamps of long-term 8-s generation. \cref{tab:ablate_masking_strategy} demonstrates the best practice of masking strategy for balancing quantitative and qualitative results is to pose the invisibility mask on control features.

\subsection{Qualitative Results}\label{sec:exp_quali}
\begin{figure}
    \centering
    \includegraphics[width=1.0\textwidth]{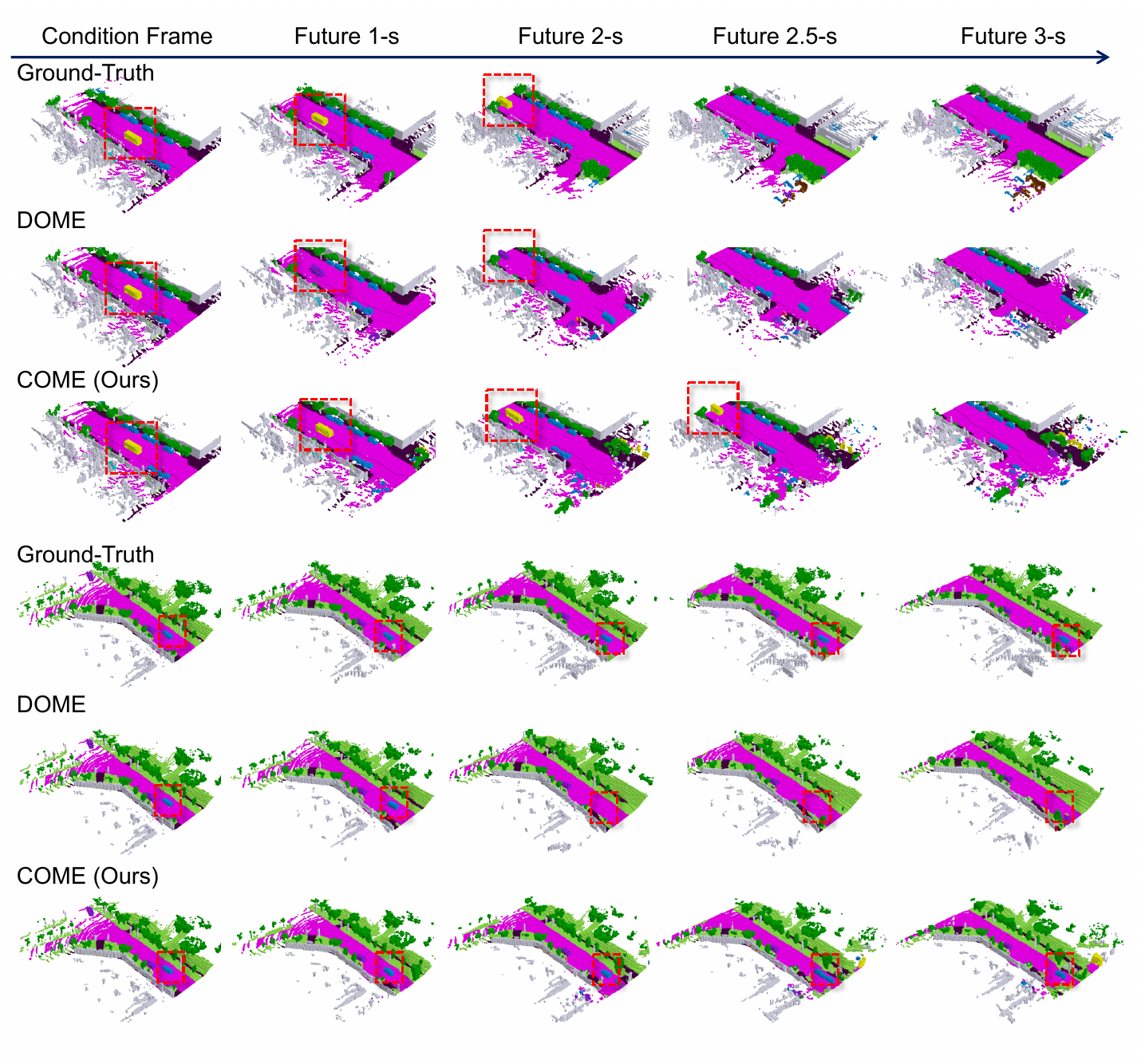}
    \caption{
        Qualitative results of 3-s 4D occupancy generation. 
        % Visualization examples demonstrate the spatiotemporal consistency of \mtd{} for generating dynamic objects. DOME loses vehicles (left image) or exhibits category jumping (right image) during generation, while our \mtd{} method maintains foreground consistency in generation.
        % \qtext{tbd}
    }
    \label{fig:come_vis_foreground}
    % \vspace{-1em}
\end{figure}

\cref{fig:come_vis_foreground} presents visualizations comparing ground-truth occupancy with predictions from DOME and our method.
We observe that DOME suffers from object category inconsistency (top example) and sudden object disappearance (bottom example).
In contrast, \mtd{} produces results with improved spatial consistency, highlighting the efficacy of the scene condition injection.
Additional qualitative results, including analyses of input variants, model architectures, and component contributions, are provided in the supplementary material (\cref{sec:more_vis}) for comprehensive evaluation.

% \qtext{
% We put more visualization results in the supplementary materials to better illustrate differences between more model variants and various inputs. They include the generation results with different driving commands and future trajectories\cref{fig:come_vis_pose_control}, with BEV layouts as input (\cref{fig:come_vis_bev_layout}), comparison between results across stages (\cref{fig:come_vis_world_model_controlnet}), comparison between different sensor inputs (\cref{fig:come_vis_sensor_input}), comparison between different masking strategies (\cref{fig:come_vis_mask_strategy}) and comparison between different model sizes (\cref{fig:come_vis_small_models}).
% }

\subsection{Ablation Study}\label{sec:exp_ab}

\begin{table}[t]
    \newcommand{\graycell}{\cellcolor{gray!30}}
    \setlength{\tabcolsep}{0.006\linewidth}
    \caption{
        \textbf{Model performance in various stages.}
        Here, ``Visible'', ``Invisible'' and ``All'' refer to voxels that are observed, unobserved, and the total set of voxels based on historical occupancy, respectively.
    }
    \label{tab:ablate_stages}
    \centering
    \small
    \begin{tabular}{l|c|ccc|ccc}
        \toprule
        \multirow{2}{*}{Model}   & \multirow{2}{*}{Input} & 
        \multicolumn{3}{c|}{mIoU (\%) $\uparrow$} & 
        \multicolumn{3}{c}{IoU (\%) $\uparrow$} \\
        & &  Visible & Invisible & All &  Visible & Invisible & All \\
        \midrule
        Stage1: World Model  & 3D-Occ (4f)     & 25.68 & 5.81 & 23.58  & 35.96 & 13.60 & 32.61 \\        
        Stage2: Scene-Centric Forecasting  & 3D-Occ (4f)       &  42.74 & 0.09 & 39.12  & 55.08 & 0.31  & 48.00 \\
        Stage3: ControlNet  & 3D-Occ (4f) & 40.06 & 5.56 & 34.23  & 51.12 & 14.95 & 44.13 \\\midrule
        % Stage1-World-Model-8-s  & 3D-Occ   & 17.77 & 4.00 &  14.76  & 29.14 & 12.91 & 24.11 \\        
        % Stage2-Scene-Centric-8-s  & 3D-Occ  & 38.23  & 0.08 &  30.35 & 54.40 & 0.18  & 39.37 \\
        % Stage3-ControlNet-8-s  & 3D-Occ & 23.16 & 4.52 & 19.07  & 36.85 & 14.45 & 29.96  \\\midrule
        Stage1: World-Model &   3D-Occ(2f),Box,Map    & 32.92 & 11.74 & 30.04  & 39.52 & 16.20 & 35.43 \\        
        Stage2: Scene-Centric Forecasting &  3D-Occ(2f)    &   41.65    &  0.08 & 37.93  & 54.50 &  0.27 & 47.18 \\
        Stage3: ControlNet  &  3D-Occ(2f),Box,Map & 42.75 & 14.85 & 39.28 & 52.78 & 20.48 &  47.16 \\
        \bottomrule
    \end{tabular}
    % \vspace{-1em}
\end{table}

\textbf{Model performances across various stages.}
In \cref{tab:ablate_stages}, we analyze the predicted occupancy across our three stages, evaluating visible, invisible, and all voxels.
Consistently, the world model performs better in previously invisible areas, while scene-centric forecasting excels in observed regions.
This validates our motivation: the generative world model exhibits strong imaginative capabilities but under-utilizes the 3D spatial consistency of the driving scene.
In contrast, scene-centric forecasting achieves high accuracy in observed areas but lacks generative flexibility, limiting its imagination applicability to novel-view synthesis under diverse trajectories.

In the third stage, we leverage the \ctln{} to combine the strengths of the two modules.
Stage three model significantly improves performance in invisible regions compared to the forecasting module.
Although the overall mIoU experiences a slight decline, the model gains imaginative capabilities and broader adaptability.
With extra BEV layout inputs, \mtd{} demonstrates further enhanced generation of invisible scenes and achieves the best overall performance.
These results showcase the effectiveness of our proposed framework.

% \qtext{
% \textbf{Effects of stages in \mtd{}.}
% \cref{tab:ablate_stages} analysis the forecasting performance under different stages. For evaluation of scene-centric forecasting, we transform the future predictions of the current coordinate to the future coordinate. Future voxels that are invisible in the current coordinate are set as free. Even if only visible region matters, scene-centric forecasting already has comparably higher performance than the world model, which outperforms stage two by 65.9\% mIoU and 47.2\% IoU. With ControlNet that injects the knowledge of scene-centric forecasting to the world model, The stage three COME outperforms the vanilla world model by 45.2\% mIoU and 35.3\% IoU. we find that the addition of the control network significantly enhances the overall metric, making the generation more reasonable.
% We also perform a simple post-fusion strategy: We use results of scene-centric forecasting of all visible regions of future voxels and use results of COME of all invisible regions of future voxels. We find this strategy further improves the mIoU metric to 41.46\% but decreases the IoU metrics. The post-fusion strategy has the risk of showing poor consistency between visible and invisible regions.
% }

\textbf{Effects of training setups.}
\cref{tab:training_ablations} presents an ablation study on different training configurations within the proposed \mtd{} framework. 
We first train \ctln{} with and without freezing the world model in the final stage, and present the results in \cref{tab:1a}. 
It can be observed that training \ctln{} while keeping the world model frozen yields significantly better results, confirming the importance of aligning with the fine-tuning strategy validated in the original \ctln{}~\cite{controlnet} paper.

In \cref{tab:1b}, we explore two model sizes for both the generative model and its corresponding \ctln{}. 
Across both settings, the introduction of \ctln{} consistently improves performance by a substantial margin. 
These gains are particularly pronounced under limited computational budgets, where mIoU increases from 7.78 to 32.00 and IoU from 18.14 to 42.03 — surpassing even the standalone world model with 1375.4 GFLOPS. 
These results demonstrate that integrating \ctln{} significantly enhances the generative capability of the model even if the generation model is small.

\begin{table}[t]
    % \vspace{-0.5em}
    \centering
    \caption{Ablations on different training setups.
    ``WM'' and ``ControlNet'' denote the world model and ControlNet in our \mtd{}, respectively.
    }
    \setlength{\tabcolsep}{0.007\linewidth}
    \begin{subtable}[t]{0.3\textwidth}
        \centering
        \caption{
            Effects of trainable parameters in the final stage.
        }
        \small
        \begin{tabular}{l|cc}
            \toprule
            Trainable modules & mIoU & IoU \\
            \midrule
            ControlNet & 34.23 & 44.13  \\
            WM\&ControlNet & 28.67 & 41.07 \\
            \bottomrule
        \end{tabular}
        \label{tab:1a}
    \end{subtable}
    \qquad
    \begin{subtable}[t]{0.6\textwidth}
        \centering
        \caption{
            Model performances under different model sizes.
        }
        \small
        \begin{tabular}{l|ccc|ccc}
            \toprule
            \multirow{2}{*}{Model} & \multicolumn{3}{c|}{Small} & \multicolumn{3}{c}{Base} \\
             & mIoU & IoU & GFLOPS & mIoU & IoU  & GFLOPS\\
            \midrule
            WM & 7.78 & 18.14  & 147.8  & 23.49 & 32.36 & 1375.4\\
            +ControlNet & 32.00 & 42.03 & 222.7 & 34.23 & 44.13 & 2066.6 \\
            \bottomrule
        \end{tabular}
        \label{tab:1b}
    \end{subtable}
    \label{tab:training_ablations}
    % \vspace{-0.5em}
\end{table}

% The scene-centric forecasting module has 27.31 M parameters and 737.76 GFLOPs. The base world model has 362.31M parameters and 1375.38 GLOPS. The base ControlNet has 158.49M parameters and 691.23 GLOPS. The base world model has 45.83M parameters and 147.85 GLOPS. The base ControlNet has 17.27M parameters and 74.86 GLOPS.

\textbf{Effects of inference setups.}
In \cref{tab:test_time_ablations}, we investigate how different inference configurations affect model performance. 
We begin by analyzing the impact of the number of denoising steps during the diffusion process, as shown in \cref{tab:2a}. 
Increasing the number of steps consistently improves generative performance. 
The results indicate a clear positive correlation between performance and the number of denoising steps, with satisfactory results achieved when performing at least 10 steps.

Next, we examine model performance using different sources of predicted occupancy and trajectories. 
In \cref{tab:2c}, we replace ground-truth occupancy with predictions from a vision-only model, BEVStereo (mIoU = 42.54), and a fusion model, EFF-Occ (mIoU = 54.08). 
The results show that stronger occupancy inputs lead to better generative outcomes.
\cref{tab:2b} shows that gradually replacing the ground-truth pose and yaw with predicted values leads to a noticeable drop in model performance. 
However, when we remove the influence of ego pose during evaluation - by aligning predicted and ground-truth occupancy to a same predicted future waypoint coordinate - the degradation is minimal, with only a 0.23 drop in mIoU and a 0.44 drop in IoU.
This phenomenon suggests that the performance degradation primarily stems from misaligned trajectories, while the generated scene quality remains relatively unaffected by trajectory errors. 
Further discussion is provided in our supplementary material (\cref{sec:discuss_eval}) .

\begin{table}[t]
% \vspace{-0.5em}
    \centering
    \caption{Ablations on various inference configurations.}\label{tab:test_time_ablations}
    \setlength{\tabcolsep}{0.007\linewidth}
    \small
    \begin{subtable}[t]{0.2\textwidth}
        \centering
        \caption{
            Effects of the denoising steps at the inference stage.
        }\label{tab:2a}
        \begin{tabular}{l|cc}
            \toprule
            \#step & mIoU & IoU \\
            \midrule
            2  & 4.41 & 14.47 \\
            5  & 4.95 & 16.17 \\
            10 & 33.42 & 44.35 \\
            20 & 34.23 & 44.13 \\
            \bottomrule
        \end{tabular}        
    \end{subtable}
    \quad
    \begin{subtable}[t]{0.33\textwidth}
        \centering
        \caption{
            Model performance when ground-truth occupancy is replaced with different occupancy generators.
            Here, ``C" and ``L" denote inputs from camera and LiDAR sensors, respectively.
            }\label{tab:2c}
        \begin{tabular}{lc|cc}
            \toprule
            Model & Input & mIoU & IoU \\
            \midrule
            BEVStereo~\cite{li2023bevstereo} & C & 22.26 & 44.07 \\
            EFFOcc~\cite{effocc} &  LC & 26.75 & 50.49 \\    
            \bottomrule
        \end{tabular}            
    \end{subtable}   
    \quad 
    \begin{subtable}[t]{0.3\textwidth}
        \centering
        \caption{
            Effects of the used trajectories.
             ``align." is the alignment operation of the ground-truth occupancies.
            }\label{tab:2b}
        \begin{tabular}{lll|cc}
            \toprule
            Pose2D & Yaw & Align. & mIoU & IoU \\
            \midrule
            GT & GT & - & 34.23 & 44.13 \\
            Pred. & GT & - & 25.90 & 35.21  \\
            Pred. & Pred. & - & 21.29 & 29.03 \\ 
            Pred. & Pred. & \checkmark & 34.00 & 43.69 \\       
            \bottomrule
        \end{tabular}        
    \end{subtable}         
% \vspace{-0.5em}
\end{table}

%% file: sec/5_conclu.tex
\section{Conclusion}\label{sec:conclusion}

We introduce \mtd{}, a framework that enhances generative occupancy world models through scene-centric forecasting control.
By explicitly decoupling ego-motion effects from scene evolution, \mtd{} first generates spatially consistent, ego-invariant control features, which are then integrated into the occupancy world model for more accurate and controllable future predictions.
Extensive experiments on the large-scale \occnus{} dataset demonstrate the state-of-the-art performances across multiple settings, validating the effectiveness of our approach for occupancy world model.

To motivate future work, we outline a few limitations based on our current
comprehension: 
(1) The introduction of control modules to the base generative model increases computational complexity. 
Although our approach achieves superior performance than the baseline with lower overhead, further optimization to reduce computations remains valuable for real-time applications.
(2) Our current multi-stage training pipeline could be streamlined. 
An end-to-end training scheme may improve efficiency while maintaining or enhancing model performance.

%% file: sec/6_appendix.tex
\appendix

% \section{World Models with End-to-End Driving}
% There are a large number of papers that integrate end-to-end driving into the design of the world model, mainly as a generator, with few practice for evaluators.

\section{Technical Appendices and Supplementary Material}

\subsection{Discussion on the Design Philosophy of External Occupancy ControlNet}

In this paper, we use a direct concatenation method for pose and BEV layout control, while we employ an external ControlNet for controlling scene-centric forecasting. We propose some explanations of the design philosophy.

Most importantly, We hope to remain the original generation capabilities of the occupancy world model unchanged. ControlNet only provides additional guidance to enhance the original world model, while direct concatenation makes the scene-centric forecasting an indispensable component for generation. This will downgrade the capability of the world model to a completion or inpainting model that inpaints future invisible voxels with the context of visible voxels. On the other hand, the model easily learns shortcut that outputs forecasting results directly on visible voxels. This shortcut learning may greatly hurt the multi-modal nature of both the generative model and the forecasting task. In contrast, ControlNet may be added to certain regions of the space or certain frames on the sequences, with user-defined masks and enjoy greater flexibility.

Moreover, poses (in the form of waypoints $[x,y,yaw]$) and bev layouts (in the form of semantic maps) are low-dimension conditions that empirically fit direct concatenation while occupancy sequences latents are high-dimension conditions that empirically fit external ControlNet. Training ControlNet usually needs smaller data scale than training the original world model. We plan to validate the effect of data scale with Occ3D-Waymo\cite{occ3d,waymo} dataset with more occupancy data for future research.

\begin{figure*}
    \centering
    \includegraphics[width=1.0\textwidth]{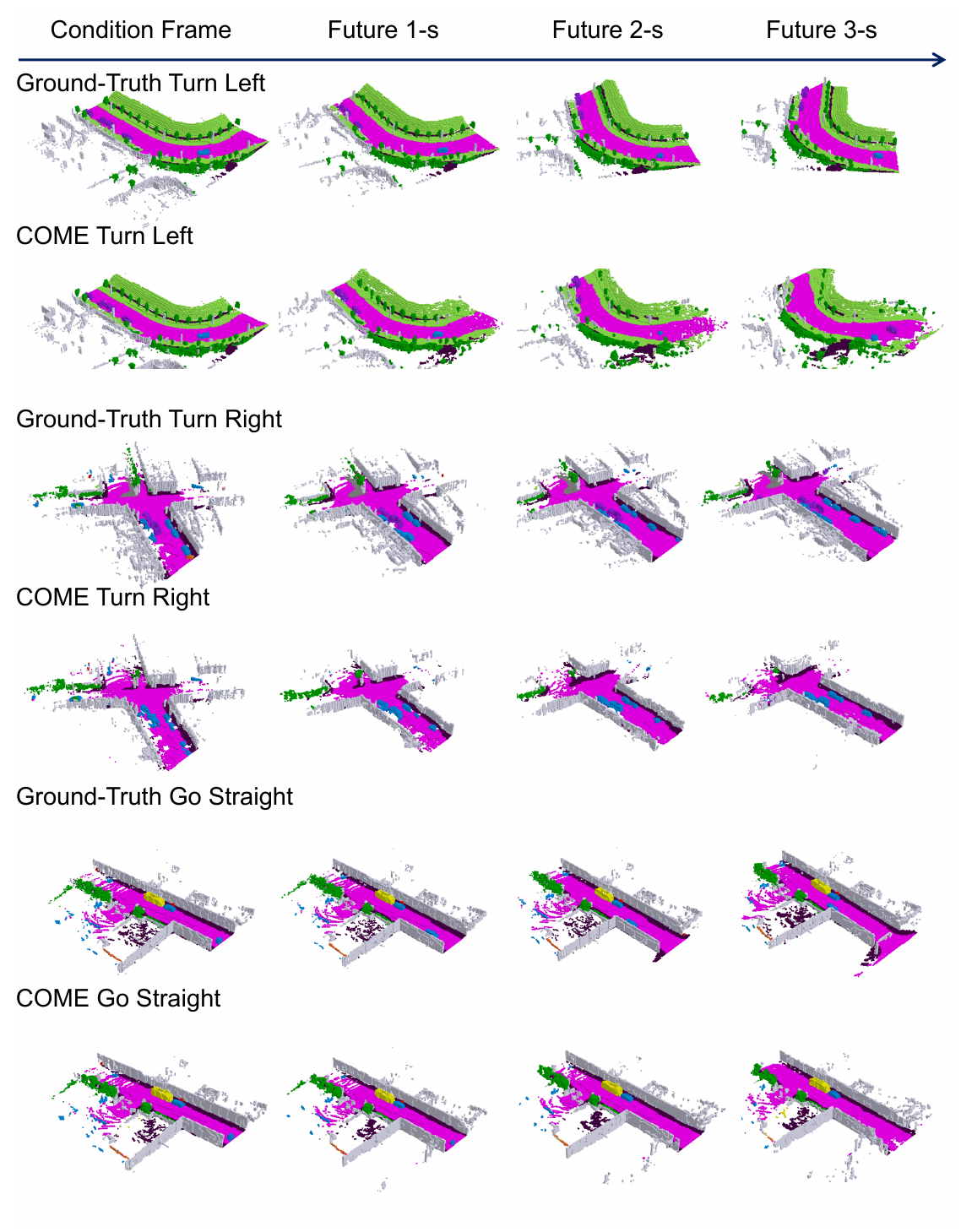}
    \caption{Visualization examples demonstrate the pose control alignment ability of \mtd{} generation. For different driving commands such as Go Straight, Turn Left and Turn Right, \mtd{} well follows the pose control and generate similar scenarios compared to ground-truth.}
    \vspace{-3mm}
    \label{fig:come_vis_pose_control}
\end{figure*}

\begin{figure*}
    \centering
    \includegraphics[width=1.0\textwidth]{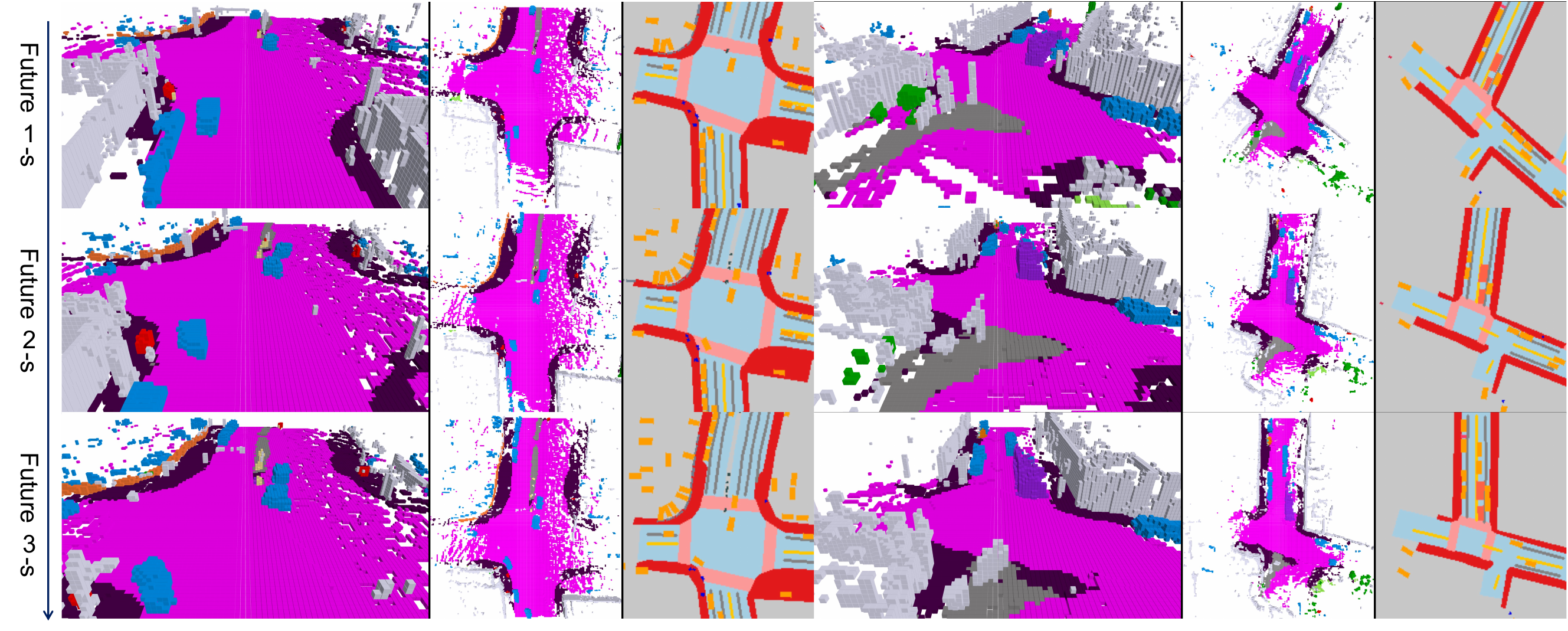}
    \caption{Visualization examples of occupancy generation with BEV layouts. \mtd{} generates occupancy sequences that well follows the BEV layout control.}
    \vspace{-3mm}
    \label{fig:come_vis_bev_layout}
\end{figure*}

\begin{figure*}
    \centering
    \includegraphics[width=1.0\textwidth]{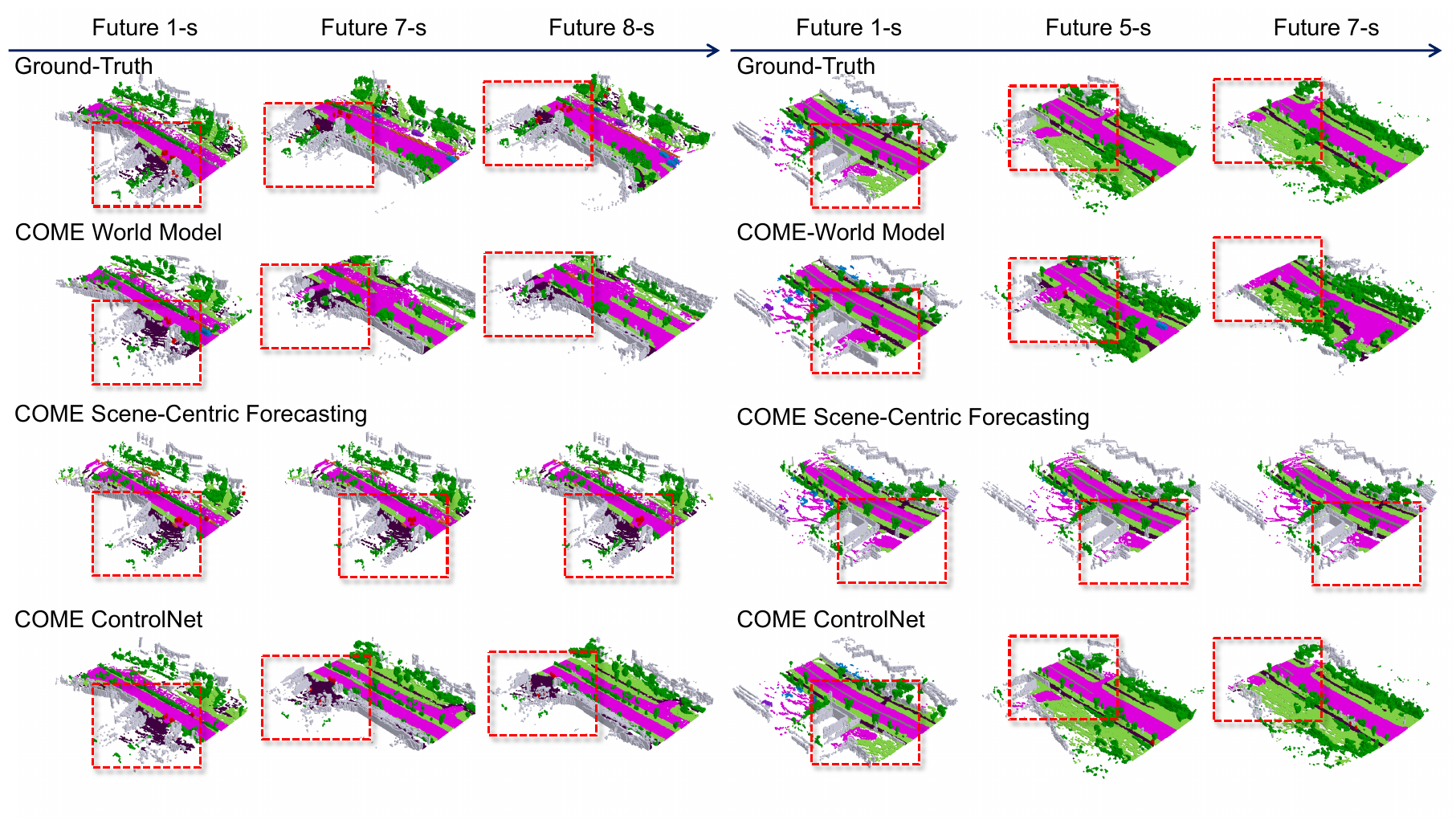}
    \caption{Visualization examples demonstrate the spatiotemporal consistency of differnet stages in \mtd{} for generating static environments. During the 8-second long generation process, \mtd{} world model mistakenly links drivable areas as intersections with the main street, but under the guidance of scene-centric forecasting, \mtd{} maintains the background consistency of road structures during generation.
    }
    \vspace{-3mm}
    \label{fig:come_vis_world_model_controlnet}
\end{figure*}

\begin{figure*}
    \centering
    \includegraphics[width=1.0\textwidth]{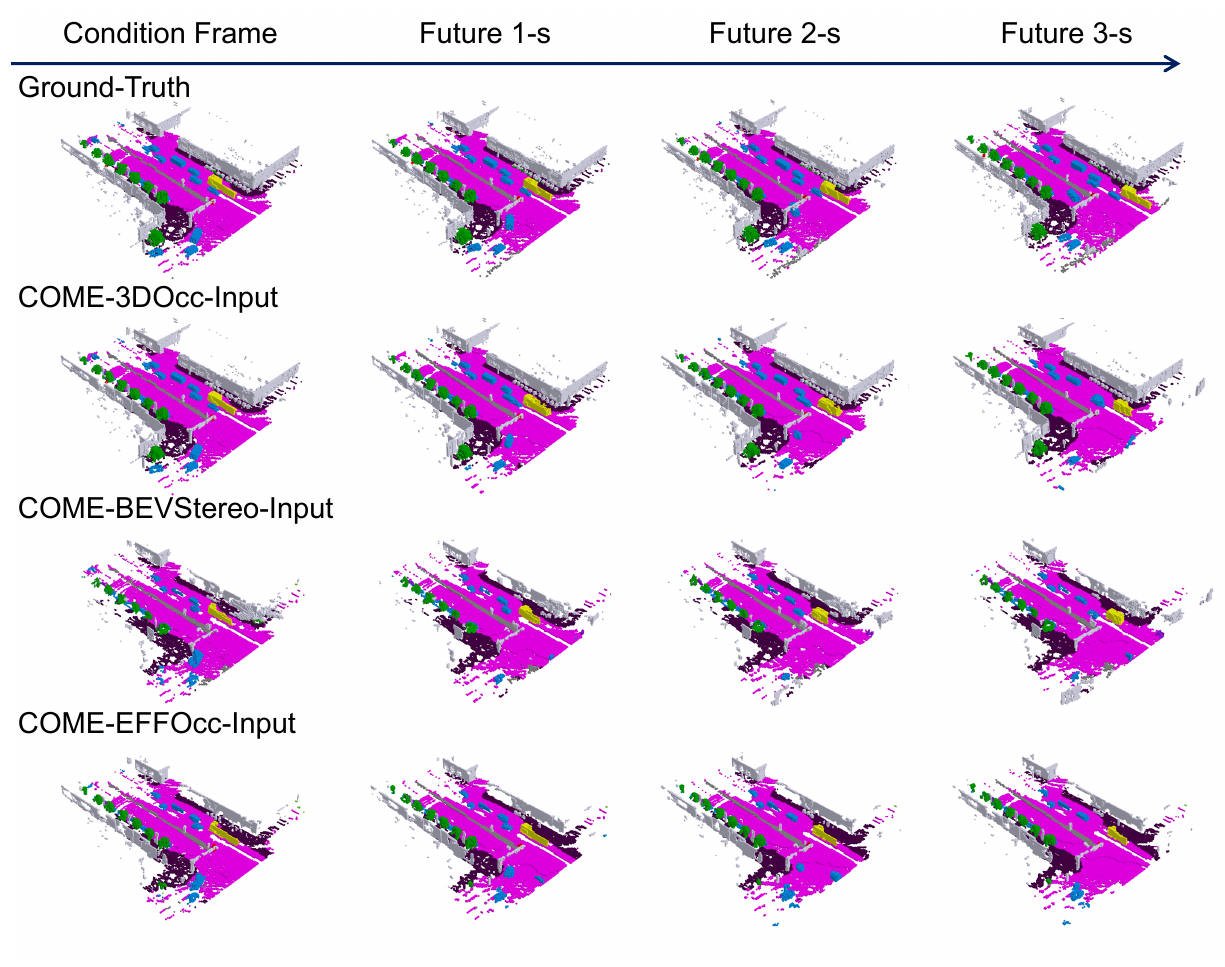}
    \caption{Visualization examples that takes 3D-Occ, vision-based BEVDet and fusion-based EFFOcc as occupancy sequences input.}
    \vspace{-3mm}
    \label{fig:come_vis_sensor_input}
\end{figure*}

\begin{figure*}
    \centering
    \includegraphics[width=1.0\textwidth]{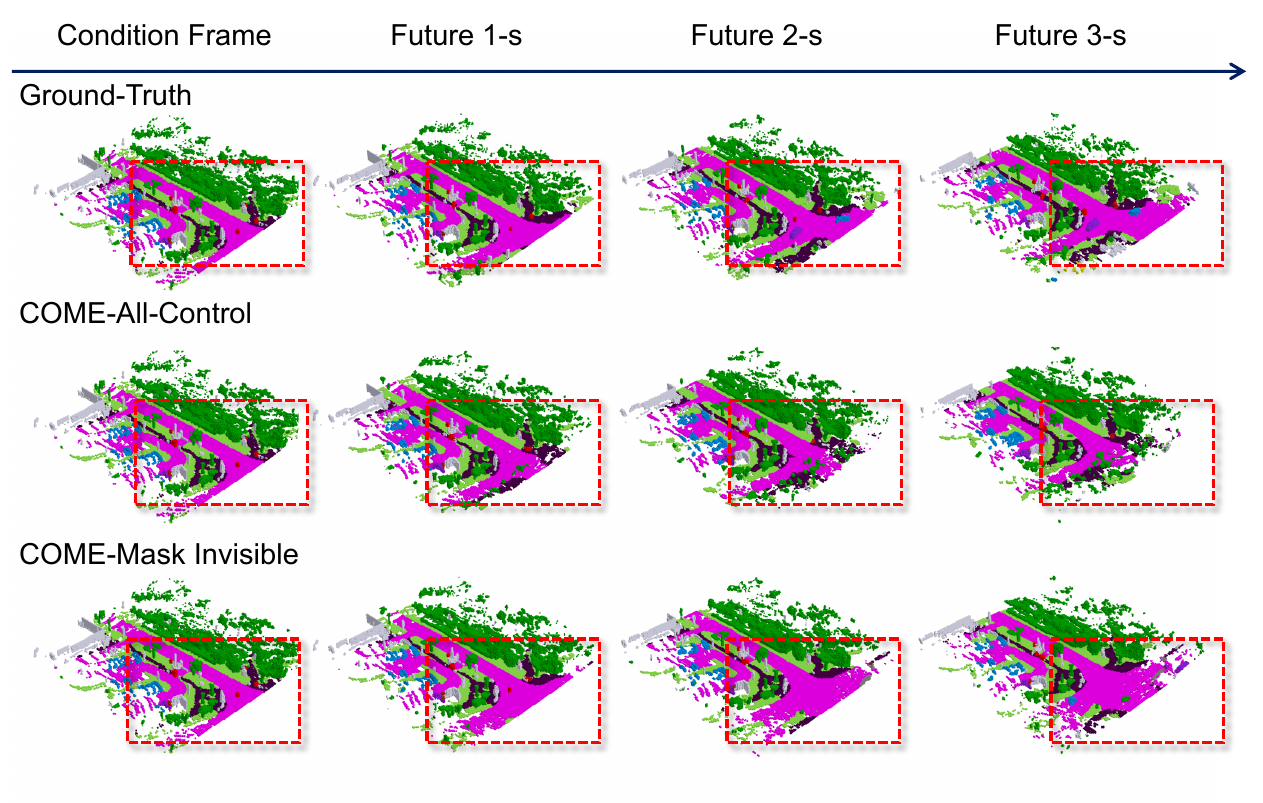}
    \caption{Visualization examples that uses different masking strategies during training and inference. Models with invisibility masks generally achieve much better qualitative results results.}
    \vspace{-3mm}
    \label{fig:come_vis_mask_strategy}
\end{figure*}

\begin{figure*}
    \centering
    \includegraphics[width=1.0\textwidth]{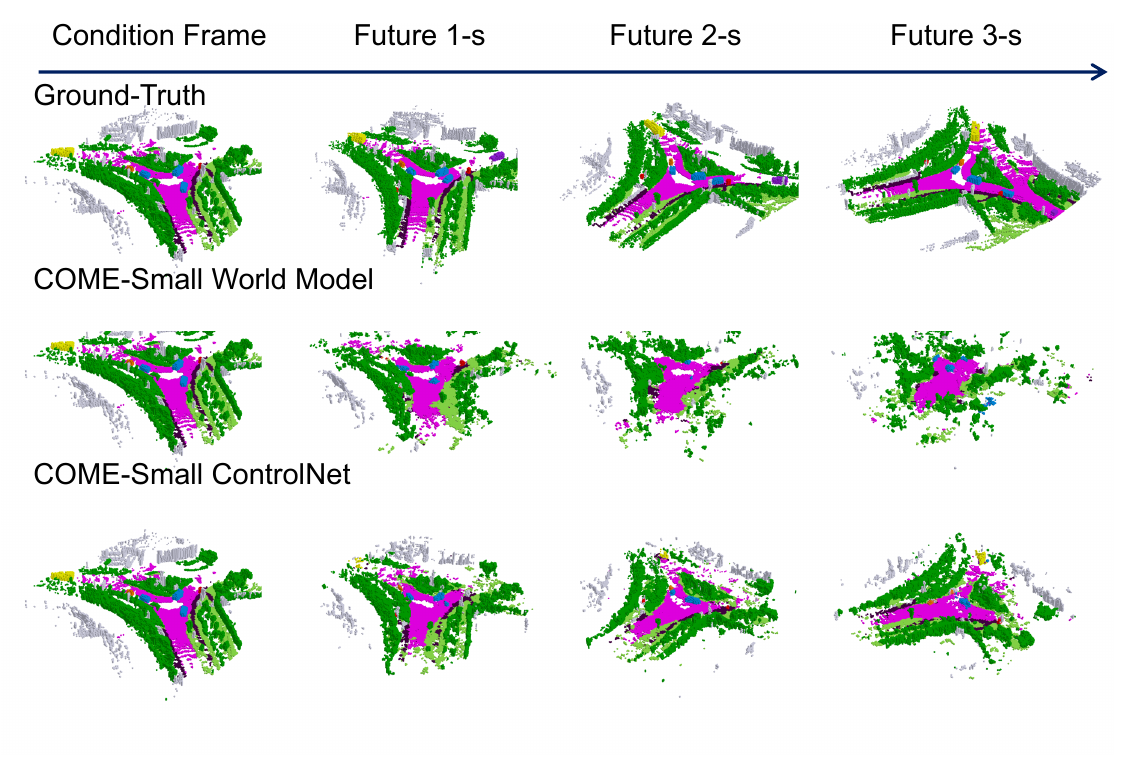}
    \caption{Visualization examples that uses small models. In this right turning example, the small model shows accurate pose control and good scene completion results.}
    \vspace{-3mm}
    \label{fig:come_vis_small_models}
\end{figure*}

\begin{figure*}
    \centering
    \includegraphics[width=1.0\textwidth]{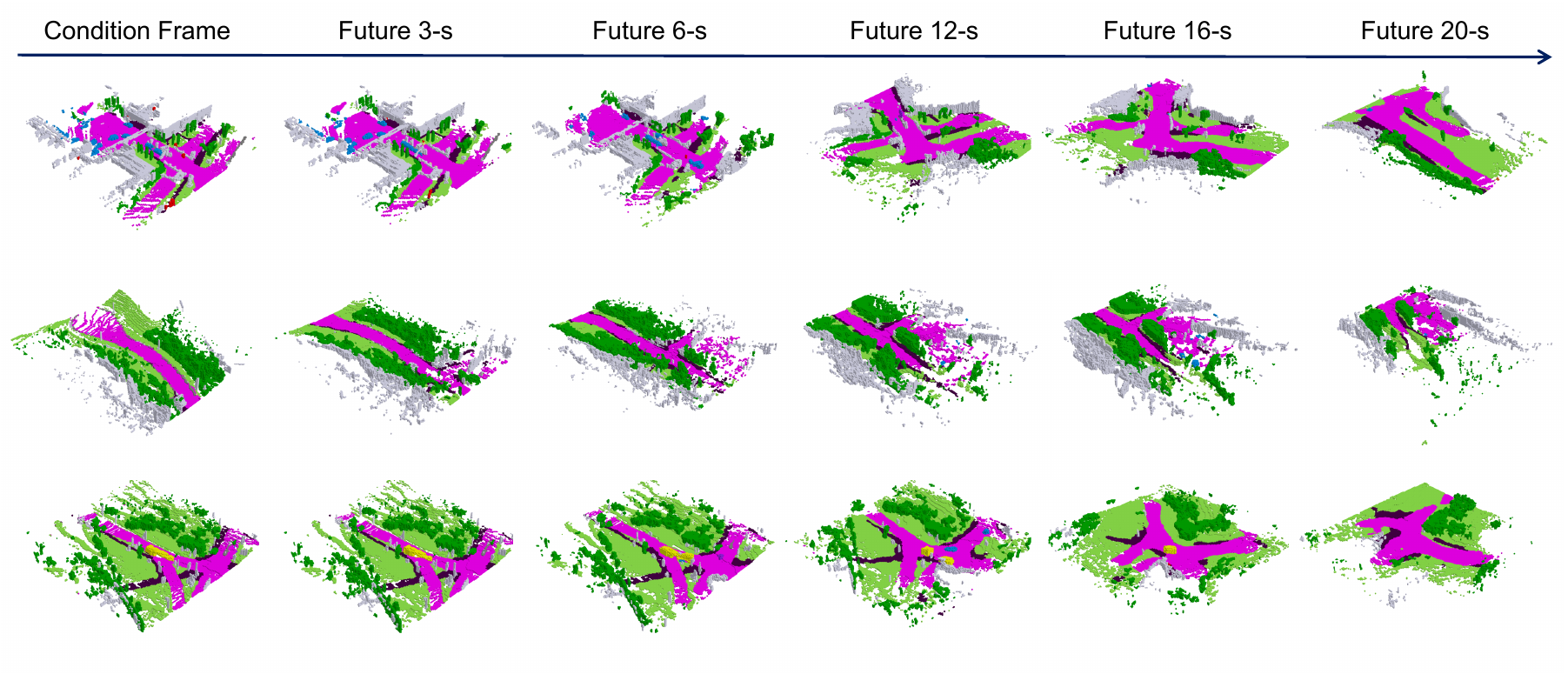}
    \caption{Visualization examples that uses repeated roll-out to generate super-long scenarios (20 second) with free trajectories.}
    \vspace{-3mm}
    \label{fig:come_vis_rollout}
\end{figure*}

\subsection{Discussion on Evaluation Metrics under End-to-end Planning Setting}\label{sec:discuss_eval}
Early autoregressive world models\cite{occworld, occllama} generates both future waypoints and future occupancy sequences at the predicted waypoints' coordinate at the same time. The evaluation between generated occupancies and ground-truth occupancies couples the similarity between the planning trajectory and the expert trajectory, and the similarity of generation. Our ablation experiment illustrates that both the translation error of the trajectory and the yaw angle error of the trajectory can greatly reduce the final IoU metric.

In order to factor out the influence of trajectory quality from the generation evaluation, we propose to reset the origin of the ground-truth occupancy according to the rotation matrix of the planning trajectory, and then evaluate the difference between the generated occupation and the ground-truth occupation. In this setting, we demonstrate that the generation performs similarly well when conditioned by the planning trajectory.

\subsection{More Implementation Details}\label{app:imp_detail}
\subsubsection{Environment Setup}
The proposed algorithm runs in the python3.9 and torch2.5.1 environment and is expected to be compatible with the torch2.x environment. The environment needs to have mmcv 2.x and mmdet3d 1.1.x installed, and it is basically the same as the environment configuration scheme of OccWorld\cite{occworld} and DOME\cite{dome}. We use AdamW optimizer for all experiments.

\subsubsection{Networks Setup}
\textbf{Continuous VAE}
We use the same VAE with DOME\cite{dome} which consists of a 2D encoder and a 3D decoder. The class embedder has the expansion of $8$. The VAE has the compression ratio of $64$. The encoder downsampling and decoder upsampling ratios are set as $[1,2,4,8]$. The base channel is set as $64$. The number of residual blocks is $2$. The attention resolution is $50$. The shapes of intermediate feature maps are $[[200, 200], [100, 100], [50, 50], [25, 25]]$.

\textbf{Scene-centric forecasting}
The class embedder has the expansion of $32$.
We use UNet with five-stage encoder and decoder. The encoder downsampling and decoder upsampling ratios are set as $[1,2,4,8,16]$ and the channel dimensions are $256,512,1024,1024,1024$ for each downsampling stride. The base channel is set as $256$. The number of temporal convolutional layers is set as $2$. We use InterpConv for upsampling operations.

\textbf{\mtd{} world model and ControlNet} We have two model sizes. By default, we use the model size similar to DiT-XL. The number of attention head is set as $12$, the hidden size is set as $768$. The number of layers (also named depth of world model) is set as $28$. We also set a smaller model. The number of attention head is set as $6$, the hidden size is set as $384$. The number of layers is set as $12$. The mlp\_ratio is set as 4. The patch size is set as 1. The topk is set as 10. For different size of world models, corresponding controlnet has half of the depth of the world model and uses the same parameters in each block. 

\textbf{Network statistics.} The following statistics is tested with the standard task of generation future three second occupancy sequences with four-frame occupancy history. The scene-centric forecasting module has 27.31 M parameters and 737.76 GFLOPs. The base world model has 362.31M parameters and 1375.38 GLOPS. The base ControlNet has 158.49M parameters and 691.23 GLOPS. The base world model has 45.83M parameters and 147.85 GLOPS. The base ControlNet has 17.27M parameters and 74.86 GLOPS.

\subsubsection{Running Parameters Setup}
\textbf{Diffusion parameters.} We use DDPM\cite{ddpm} as the default diffuser. By default, we use 1000 denoising steps for training and 20 denoising steps for inference. We also find in the ablation that 10 denoising steps for inference only very slightly drops the final performance. The guidance scale is set as $7.5$. If the historical frame number is 4, the conditional frames in training may be $[],[0],[0,1],[0,1,2],[0,1,2,3]$ and the conditional frames in inference is $[0,1,2,3]$. The possibility of using pose as condition in training is set as $0.9$.

\subsubsection{Optional Inputs}
\textbf{Sensor Inputs.} For vision track, we use the officially released checkpoint of BEVDet\cite{bevdet}, BEVStereo, with Swin Transformer\cite{swin} base and image size $512\times1408$ as input and 1 historical frames, achieves 3D occupancy prediction mIou of $42.45$. For LiDAR-camera fusion track, we use the officially released checkpoint of EFFOcc\cite{effocc}. EFFOcc, with Swin Transformer\cite{swin} base and image size $512\times1408$ as input, achieves 3D occupancy prediction mIou of $54.08$. As 3D occupancy models are trained with camera mask on nuScenes dataset, we also use camera mask during occupancy generation evalaution.

\textbf{Planning trajectory Input}
\mtd{} does not strongly bind the world model to planning, instead, \mtd{} is controlled by the ego future trajectory as input. To compare with other world models with planning, we train a unsupervised simple imitation learning planning framework built upon BEV-Planner\cite{bev-planner}. The planning module takes multi-view images as sensor input and outputs six waypoints including 2d translation and relative yaw angle compared to the current frame. The only supervision is the expert trajectory with ground-truth yaw angles. The planning module has an 3-s average L2 error of $0.48m$ under BEV-Planner open-loop metric. 

The vast majority of planning algorithms are centered around the current coordinate of the ego vehicle, without including past or future perspectives, so more complicated planning modules can be placed as downstream or a parallel heads which integrates with the scene-centric forecasting module. In this way, planning trajectories can be integrated as control conditions into the generative model and theControlNet. We leave better planning modules considering scene-centric forecasting as future research.

\textbf{BEV Layout Input} The pre-processing for BEV layouts are the same as UniScene\cite{uniscene}. 3D bounding boxes of annotated objects are splatted to the BEV plane. The static layouts are computed with polygons of the high-definition maps. 

% In order to compare with other methods that generate trajectories and occupy trajectories at the same time, in addition to inputting the truth trajectories, we can also realize the self-driving decision under the fixed perspective of the current frame coordinate system.

% We find that in the past, only 3D-Occ was used as the end-to-end decision network for the context of the contextual information, which led to the suboptimal open-loop decision-making effect, because the image itself has rich semantic and textural information, so the original image needs to be used for decision-making.

\subsection{More Quantitative Results}\label{sec:more_results}

\begin{table}[t]
    \newcommand{\graycell}{\cellcolor{gray!30}}
    \setlength{\tabcolsep}{0.007\linewidth}
    \caption{\textbf{Long-term 4D occupancy generation performance.} 
    Ground-truth 3D occupancy and trajectories are used as inputs. 
    The best results are highlighted in bold. 
    ``Avg." denotes the average performance over the 1 second to 8 second horizon.
    }       
    \label{tab:compare_long}
    \centering
    \small
    \begin{tabular}{l|cccccccc|c}
        \toprule
        \multirow{2}{*}{Method} &             
        \multicolumn{9}{c}{mIoU (\%) $\uparrow$} \\
        &  1s & 2s & 3s & 4s & 5s & 6s & 7s & 8s & \graycell  Avg. \\
        \midrule
        DOME~\cite{dome}    & 30.10 & 21.35  & 17.36 & 14.86 & 12.61    & 11.03 & 10.00 & 9.34  &\graycell 15.83  \\
        \mtd{ (Ours)}      &   \textbf{33.78}  &  \textbf{24.57}    &  \textbf{21.35} & \textbf{18.25} & \textbf{15.84}   & \textbf{13.85} & \textbf{12.99} & \textbf{11.96} &\graycell {\textbf{19.07}} \\ \midrule            
        \multirow{2}{*}{Method} & 
        \multicolumn{9}{c}{IoU (\%) $\uparrow$}  \\
        & 1s & 2s & 3s & 4s & 5s & 6s & 7s & 8s & \graycell Avg. \\ \midrule
        DOME~\cite{dome}       & 39.04 & 31.20 & 27.14 &  24.73 & 22.32  & 20.28 & 19.05 & 17.97 & \graycell  25.21 \\
        \mtd{ (Ours)}     &   \textbf{44.20}  &  \textbf{36.25}    &  \textbf{32.86} & \textbf{30.03} & \textbf{26.93}   & \textbf{24.70} & \textbf{23.30} & \textbf{21.44} &\graycell {\textbf{29.96}}  \\ 
        \bottomrule
    \end{tabular}
\end{table}

\subsubsection{Long-term Occupancy Generation}
The ability to generate long-term predictions is crucial for world models, particularly in the context of autonomous driving. 
To evaluate this capability, we extend the prediction horizon from 3 seconds to 8 seconds and present the results in \cref{tab:compare_long}. 
Our method, \mtd{}, consistently outperforms baselines across all timestamps and on average, for both mIoU and IoU metrics. 
Specifically, \mtd{} achieves average mIoU and IoU scores of 19.07 and 29.96, surpassing DOME by 20.5\% and 18.8\% respectively.

\subsubsection{Generation with Different Masking Strategies}
\cref{tab:ablate_masking_strategy} demonstrates how masking strategies affect the final quantitative results. We find that model without any masking performs the best mIoU and IoU metrics, but it has a strong tendency to generate invisible areas as free. On the other hand, the model with masked control has much better visualization performance, but the quantitative results are lower than model without masking. In the main paper, we use model with masked control by default.

We find the possible reason for the formation of a trend that opposes quantitative results and qualitative results, is that the deductions for incorrect generation of invisible areas (generated content that does not match the true value) are higher for non-generation of invisible areas (where all areas that need to be generated are left blank) under the existing metrics.

We also try with different masking strategies. We test perform masks on conditions before \mtd{} ControlNet, but the qualitative results are also as poor as models without masking, so we do not use this strategy. We test with random dropout of masking in training stage, this operation helps better visualization quality but decreases the quantitative results. Finally, we find the best practice to balance quantitative results and qualitative results, that is to use a fixed invisibility mask on control features after \mtd{} ControlNet both in training and inference stages. As a result, we report this model and its variants in the main paper.

\begin{table}[t]{}
        \centering
        \newcommand{\graycell}{\cellcolor{gray!30}}
        \caption{Results on \mtd{} with different control masking strategies.}
        \begin{tabular}{l|cccc|cccc}
        \toprule            
        \multirow{2}{*}{Masking Strategy}  &
        \multicolumn{4}{c|}{mIoU (\%) $\uparrow$} & 
        \multicolumn{4}{c}{IoU (\%) $\uparrow$} \\
         &  1s & 2s & 3s & \graycell  Avg. & 1s & 2s & 3s & \graycell Avg. \\\midrule
            No Mask & 38.33  & 35.05 & 29.90 & \graycell {36.19}  & 51.67 & 45.32 & 40.90 &\graycell {45.97} \\
            Mask Condition & 43.25 & 34.05 & 28.82 & \graycell35.37 & 50.86& 43.78 & 39.05 & \graycell44.56 \\
            Random Dropout & 40.32 & 30.55 & 36.69 & \graycell 31.97 & 48.78 & 41.78 & 25.04 & \graycell42.42 \\
            Mask Control & 42.75 & 32.97 & 26.98 & \graycell 34.23 & 50.57 & 43.47 & 38.36 & \graycell 44.13 \\
            \bottomrule
        \end{tabular}
        \label{tab:ablate_masking_strategy}
    \end{table}

\subsection{More Visualizations}\label{sec:more_vis}

We show more visualization results including the generation results with different driving commands and future trajectories\cref{fig:come_vis_pose_control}, with BEV layouts as input (\cref{fig:come_vis_bev_layout}), comparison between results across stages (\cref{fig:come_vis_world_model_controlnet}), comparison between different sensor inputs (\cref{fig:come_vis_sensor_input}), comparison between different masking strategies (\cref{fig:come_vis_mask_strategy}), visualization with small models (\cref{fig:come_vis_small_models}), super-long occupancy video generation (\cref{fig:come_vis_rollout}).

\cref{fig:come_vis_pose_control} demonstrates that \mtd{} can well align occupancy generation results with different driving commands (turn left, turn right, go straight) and pose control. The accurate pose control credits to the explicit transformation modelling guidance from scene-centric forecasting.

\cref{fig:come_vis_bev_layout} demonstrates that \mtd{} can well align occupancy generation results with BEV layouts as condition.

\cref{fig:come_vis_world_model_controlnet} demonstrates an example of how scene-centric forecasting helps ego-centric generation. With \mtd{} world model ego-centric generation alone, after a long generation time, a new path appears at the roadside where there is originally a separating strip, forming an intersection, which is inconsistent with previous observations. The scene-centric forecasting easily learns the static nature of the scenario. With scene-centric forecasting as control, \mtd{} ControlNet generates correct road structures and avoid the inconsistency caused by \mtd{} world model.

\cref{fig:come_vis_sensor_input} demonstrates that the generation from sensor inputs are similarly well for the same scene. When the 3D occupancy quality is better, the generation results are also better.

\cref{fig:come_vis_mask_strategy} compares the generation results with different masking strategies. If we add full-space control regions, the model has an increasing tendency of generation invisible areas as free. If we mask control features on invisible areas (\mtd{}-Mask Invisible), the generation results are more satisfactory on most cases. We find that the performance of the visualization is not always consistent with the quality of the metric, which may be due to the fact that the metric was too restrictive. Specifically, when we mask the control of unseen areas, it is very reasonable to have free generation of unseen areas, but the reported metrics decrease because it is very likely that free generation is not the same as the truth value. If we exert control over the whole world, there is a higher tendency to set the unseen area to free, which in turn improves the quantitative results.

\cref{fig:come_vis_small_models} shows an example that the smaller model achieves equally good generation results with ControlNet. The ego vehicle is turning at a large angle. The small world model has limited generation quality. However, with ControlNet, the small \mtd{} demonstrates better pose control and satisfactory generation results. 

\cref{fig:come_vis_rollout} shows some examples that the model roll-outs several times to generate super-long videos. This roll-out mechanism is similar to DOME\cite{dome} except that we use \mtd{} ControlNet for the first roll and use \mtd{} world model without ControlNet for the next rolls. This shows the fleixibility of our framework with and without ControlNet guidance at the same time.